# AdjointBackMap: Reconstructing Effective Decision Hypersurfaces from CNN Layers Using Adjoint Operators

Qing Wan*[1] and Yoonsuck Choe†[1]

[1]Department of Computer Science and Engineering
Texas A & M University, College Station, TX, USA, 77843

**Abstract**

There are several effective methods in explaining the inner workings of convolutional neural networks (CNNs). However, in general, finding the inverse of the function performed by CNNs as a whole is an ill-posed problem. In this paper, we propose a method based on adjoint operators to reconstruct, given an arbitrary unit in the CNN (except for the first convolutional layer), its effective hypersurface in the input space that replicates that unit's decision surface conditioned on a particular input image. Our results show that the hypersurface reconstructed this way, when multiplied by the original input image, would give nearly the exact output value of that unit. We find that the CNN unit's decision surface is largely conditioned on the input, and this may explain why adversarial inputs can effectively deceive CNNs.

## 1 Introduction

Convolutional Neural Network (CNN) has seen great success in computer vision (CV). It is based on the receptive fields found in biological vision [1], implemented by convolution [2], and trained by backpropagation [3]. Since 2009, hardware advancements like GPUs [4] have enabled the fast computation of CNNs, securing their dominance in the machine learning field.

CNN gradually learns more complex visual features through its multi-layer convolutional architecture. Deconvolution [5] can illustrate what represent in each layer: edge features at lower layers and object features at higher layers. CAM [6] and its successor, Grad-CAM, [7] can highlight specific features in the input image that contributes most to the decision made by CNN. Especially, Grad-CAM, with guided backpropagation, locates input features that directly influenced the output [8], indicating which specific part the CNN used.

However, it is also known that CNNs are vulnerable to adversarial attacks. Usually, these adversarial patterns cannot deceive humans. For example, [9] showed an adversarial pattern that makes a CNN mistakenly recognize a panda as a gibbon by adding a low level of noise while the original panda image and its contaminated version are visually identical to a human. Also, recent research showed that a physical adversarial attack (a T-shirt worn by a human with an adversarial pattern) could deceive a CNN-based object detection model, YOLOv2. Thus, the way CNN processes visual input may be profoundly different from those of humans.

These differences leave us with a fundamental question: how does CNN make its decision? Furthermore, what does the decision boundary for a feature map unit in CNN look like? However, CNN employs a cascade of convolutional layers with nonlinear activations like ReLU [10]. A kernel in a specific layer only plays its hyperplane role on that layer. It does not give us much intuition into its role played in the input space if we take the kernel out and visualize it directly, especially as we go high in the layer. Inverting the function performed by a feature map may be a better way. However, the inverse of a CNN is an ill-posed problem because bijection does not exist in the mapping from input to CNN output.

In this paper, we introduce a novel algorithm (AdjointBackMap) that aims at precisely reconstructing an effective hyperplane of a CNN. We circumvent the difficulty of finding the inverse by mapping every convolution kernel (in higher layers) or weight vector (in the fully connected [FC] layer) back to the input image space with the help of adjoint operators [11]. As long as two necessary conditions are satisfied (see the Algorithm section), our theory guarantees that,

1. In any convolutional layer, AdjointBackMap maps any convolutional kernel from that layer back to the original input image space with RM4 (Eq.6) to RM1 (Eq.9) Reconstruction Modes (RMs);

*frankqingwan@gmail.com
†choe@tamu.edu

2. In the fully connected (FC) layers, AdjointBackMap maps the final decision weights from the top layer back to the input image space through (Eq.10) ($RM_0$) reconstruction mode;

3. Any effective hypersurface reconstructed above by AdjointBackMap is mathematically precise;

4. Any single hypersurface reconstructed by AdjointBackMap represents the whole decision process from that input to a unit in the feature map or the output value of the FC layer output.

From our results, we learn that:

1. Although a kernel convolves on every receptive field location, it may not utilize information from all those locations;

2. All effective hypersurfaces mapped by AdjointBackMap on high-level kernels are not human-recognizable patterns in the input image space. Since these hypersurfaces decide a high-level feature map or FC output, they suggest that CNN's decision is not like that of human vision;

3. Any reconstructed effective hypersurface is conditioned on the current input image $\boldsymbol{x}$. This explains why adversarial examples are possible because two effective decision hypersurfaces of a CNN corresponding to two human-indistinguishable images can be very different from each other.

Based on these results, we can say that the way CNN's predict is significantly different from how human vision works. Effective hypersurfaces of CNN are brittle due to input dependence, and adversarial examples can take advantage of this property.

## 2 Related Work

Understanding the decision process of a CNN is an active research direction towards explainable AI. Attempts to invert the CNN model include works like [12, 13]. These methods invert a learned feature map in CNN back to the input space to visualize parts of the input image contributing to the feature map's output. A slightly different approach to the above is to explain a CNN with gradient or saliency [6, 14, 7, 15].

Perturbation is another path to estimate feature importance inside a CNN. It treats a deep learning model as a black box and observes how prediction changes when input varies [16, 17, 18, 19, 20].

What seems to be missing from the works above is, an input feature is identified through these methods from a given CNN, when that specific input feature pattern is fed back into the same CNN, the output classification should be the same. Furthermore, the actual output layer activation values should be similar to the activation due to the original input image. For example, consider the example is illustrated in Figure 1, based on [21]. A guided-bp pattern (right), fed into

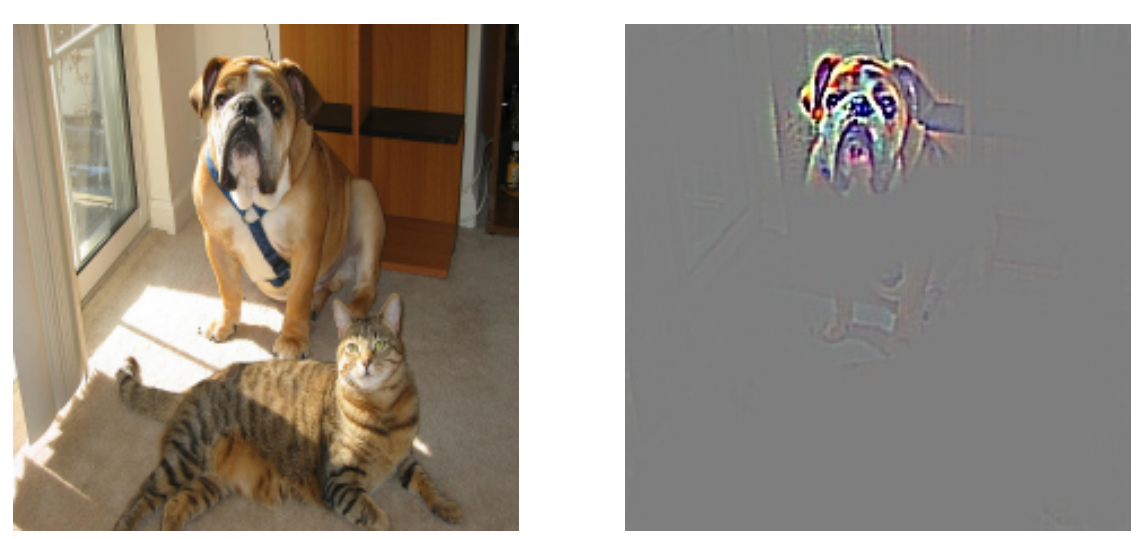

(a) Original dog&cat image, VGG16 recognizes as Boxer (n02108089)

(b) The guided-bp pattern, VGG16 recognizes as Balloon (n02782093)

Figure 1: The guided-bp pattern (right) fed into VGG16, gives a different prediction. If VGG16 did use the highlighted area on guided-bp in its "Boxer" prediction, we would expect to see the same "Boxer" when feeding this highlighted area. However, VGG16 predicts it as a "Balloon".

the same VGG16, gives a misclassification. This implies it may not be appropriate to isolate some feature from the original image and claim it as the main factor in CNN's prediction. Considering this, we built AdjointBackMap so that the inner product between the input image and our reconstructed hypersurface is numerically equal to the feature map's activation value or the FC layer unit.

Our experiment found that the AdjointBackMap reconstruction mode $RM_0$ shares some similarities with the saliency method. However, the theoretical origins of these two are different, and the saliency method does not reproduce the precise activation value.

# 3 Contributions

Our contributions are as follows: (1) Utilization of the concept of Adjoint Operators in the analysis of decision process inside a CNN; (2) Work directly on conv kernels instead of feature maps from the dual of input image space; (3) Precisely recover a complex decision process leading to the activation value of a feature map or an FC-layer output inside a CNN by visualizing a reconstructed effective decision hypersurface; (4) Discovery that CNN's decision hypersurface is largely conditioned on the current input; (5) Visualize variations in CNN's effective hypersurfaces under adversarial attack; (6) Visual evidence from effective hypersurfaces shows CNN's decision process is significantly different from our human vision.

# 4 Model

## 4.1 Notation

We will use the following mathematical notations. (1) $\circledast$ denotes convolution. (2) $\theta_{\mathcal{X}}$ denotes the origin of the vector space $\mathcal{X}$; (3) $\langle \boldsymbol{x} \mid \boldsymbol{y} \rangle$ denotes the inner product of $\boldsymbol{x}, \boldsymbol{y} \in \mathcal{X}$; (4) $\mathcal{X}^*$ denotes the algebraic dual of $\mathcal{X}$, i.e. the space of all linear functionals on $\mathcal{X}$; (5) $\langle x, x^* \rangle$ denotes the value of a linear functional $x^* \in \mathcal{X}^*$ at $x \in \mathcal{X}$; (6) $B(\mathcal{X}, \mathcal{Y})$ denotes the space of all bounded linear operators from $\mathcal{X}$ to $\mathcal{Y}$.

## 4.2 Theory

Convolution distinguishes CNNs from other artificial neural networks and gives them power. The convolution process acts on the in-channel feature map with receptive-field-sized weights (the kernel). Below, we will consider convolution from an algebraic dual perspective.

With the help of Frechet differential $\delta \boldsymbol{F}(\boldsymbol{x}; \boldsymbol{h})$ [22, 23] on increment $\boldsymbol{h}$, that

$$\frac{\|\boldsymbol{F}(\boldsymbol{x}+\boldsymbol{h}) - \boldsymbol{F}(\boldsymbol{x}) - \delta \boldsymbol{F}(\boldsymbol{x}; \boldsymbol{h})\|_{\mathcal{Y}}}{\|\boldsymbol{h}\|_{\mathcal{X}}} \to 0, \|\boldsymbol{h}\|_{\mathcal{X}} \to 0, \tag{1}$$

where $\boldsymbol{F} : \mathcal{X} \to \mathcal{Y}$ be an operator from an open domain $\mathcal{D}$ of a normed space $\mathcal{X}$ to a normed space $\mathcal{Y}$, we are able to approximate a convolution inside a CNN layer. Suppose $\boldsymbol{F}$ denotes a forward path from a small input image $\boldsymbol{x} \in \mathcal{X}$ to an in-channel $r_1 \times r_2$ receptive field represented by $\mathcal{Y}$ on the feature map of a CNN which has no bias unit, i.e. $\boldsymbol{F}(\boldsymbol{\theta}_{\mathcal{X}}) = \boldsymbol{\theta}_{\mathcal{Y}}$, then

$$\|\boldsymbol{F}(\boldsymbol{x}) - \boldsymbol{F}(\boldsymbol{\theta}_{\mathcal{X}}) - \boldsymbol{J}_{\boldsymbol{F}}(\boldsymbol{\theta}_{\mathcal{X}})\boldsymbol{x}\|_{\mathcal{Y}} = \|\boldsymbol{F}(\boldsymbol{x}) - \boldsymbol{J}_{\boldsymbol{F}}(\boldsymbol{\theta}_{\mathcal{X}})\boldsymbol{x}\|_{\mathcal{Y}} \sim o(\|\boldsymbol{x}\|_{\mathcal{X}}), \tag{2}$$

where $\boldsymbol{J}_{\boldsymbol{F}} : \mathcal{X} \to B(\mathcal{X}, \mathcal{Y})$ denotes the Jacobian Operator, and $o(\|\boldsymbol{x}\|_{\mathcal{X}})$ denotes higher order of $\|\boldsymbol{x}\|_{\mathcal{X}}$. Therefore, we write a kernel $\boldsymbol{w}_{r_1 \times r_2}$ that convolves on $\boldsymbol{F}(\boldsymbol{x}_0)$ ($\boldsymbol{x}_0$ is a fixed nonzero image) as,

$$c = \boldsymbol{F}(\boldsymbol{x}_0) \circledast \boldsymbol{w}_{r_1 \times r_2} = \langle \boldsymbol{F}(\boldsymbol{x}_0), \boldsymbol{w}_{r_1 \times r_2} \rangle \approx \langle \boldsymbol{J}_{\boldsymbol{F}}(\boldsymbol{z}(\boldsymbol{x}_0))\boldsymbol{x}_0, \boldsymbol{w}_{r_1 \times r_2} \rangle, \tag{3}$$

where $c \in \mathbb{R}$ represents a unit in the out-channel feature map; $\boldsymbol{J}_{\boldsymbol{F}}(\boldsymbol{z}(\boldsymbol{x}_0)) \in B(\mathcal{X}, \mathcal{Y})$ if $\exists \boldsymbol{z}$, $\boldsymbol{z}(\boldsymbol{x}_0) \in \mathcal{X}$ turns the last "≈" to "=". The proof of Section B (Appendix) will show a replacement that $\boldsymbol{z}(\boldsymbol{x}_0) = k\boldsymbol{x}_0$ for $\forall k \in \mathbb{R}^+$ achieves this "=" and also imply that $\boldsymbol{x}_0$ does not have to be small when a CNN is activated by either ReLU or Leaky ReLU.

Considering the input image space $\mathcal{X}$ as a Hilbert space with its norm induced on the inner product defined element-wise as:

$$\langle \boldsymbol{x}_{H \times W \times C} \mid \boldsymbol{y}_{H \times W \times C} \rangle = \sum_{i=1}^{H} \sum_{j=1}^{W} \sum_{k=1}^{C} x_{i,j,k} y_{i,j,k}, \tag{4}$$

by Adjoint Operator $\boldsymbol{J}^*_{\boldsymbol{F}}(\boldsymbol{z}(\boldsymbol{x}_0))$ of the Jacobian, we have

$$c = \langle \boldsymbol{J}_{\boldsymbol{F}}(\boldsymbol{z}(\boldsymbol{x}_0))\boldsymbol{x}_0, \boldsymbol{w}_{r_1 \times r_2} \rangle = \langle \boldsymbol{x}_0 \mid \boldsymbol{J}^*_{\boldsymbol{F}}(\boldsymbol{z}(\boldsymbol{x}_0))\boldsymbol{w}_{r_1 \times r_2} \rangle = \langle \boldsymbol{x}_0 \mid (\boldsymbol{J}_{\boldsymbol{F}}(\boldsymbol{z}(\boldsymbol{x}_0)))^T \boldsymbol{w}_{r_1 \times r_2} \rangle. \tag{5}$$

Here $\boldsymbol{w}_{r_1 \times r_2} \in \mathcal{Y}^*$ and $\boldsymbol{J}^*_{\boldsymbol{F}}(\boldsymbol{z}(\boldsymbol{x}_0)) \in B(\mathcal{Y}^*, \mathcal{X}^*)$.

From Riesz Representation theorem [24] (a hyperplane: $\{\boldsymbol{x} \in \mathcal{X} \mid \langle \boldsymbol{x} \mid \boldsymbol{J}^*_{\boldsymbol{F}}(\boldsymbol{z}(\boldsymbol{x}_0))\boldsymbol{w}_{r_1 \times r_2} \rangle = c\}$), we know $\mathcal{X}^*$ is $\mathcal{X}$ itself. So the adjoint operator will map a convolutional kernel sitting on any layer of a CNN (except the first convolutional layer) back to the input image space that serves as an effective hypersurface representing all decision hyperplanes forward from the input to the out-channel feature map from that layer, which enables our visualization.

We mention two things: (1) The approximation in Eq.3 is not linear since $\exists \boldsymbol{x}, \boldsymbol{y} \in \mathcal{X}$, such that, $\boldsymbol{J}_{\boldsymbol{F}}(\boldsymbol{z}(\alpha\boldsymbol{x} + \beta\boldsymbol{y}))(\alpha\boldsymbol{x} + \beta\boldsymbol{y}) \neq \alpha \boldsymbol{J}_{\boldsymbol{F}}(\boldsymbol{z}(\boldsymbol{x}))\boldsymbol{x} + \beta \boldsymbol{J}_{\boldsymbol{F}}(\boldsymbol{z}(\boldsymbol{y}))\boldsymbol{y}$ for $\alpha, \beta$ scales; (2) $\boldsymbol{J}^*_{\boldsymbol{F}}(\boldsymbol{z}(\boldsymbol{x}_0))\boldsymbol{w}_{r_1 \times r_2}$ is in the dual space, $\mathcal{X}^*$, of $\mathcal{X}$ (Figure 2). We will view this $\boldsymbol{J}^*_{\boldsymbol{F}}(\boldsymbol{z}(\boldsymbol{x}_0))\boldsymbol{w}_{r_1 \times r_2}$ in the dual space instead of in the input image space if the Riesz Representation is not applied. In other words, Riesz Representation frees us from visualizing two distinct spaces.

### 4.3 Algorithm

Generally, we deploy our AdjointBackMap on two kinds of layers in a CNN (Figure 2), (1) We map a filter from any convolutional layer (except for the first layer, from which a kernel has belonged to the dual of input) back to input image space and visualize it as an effective hypersurface that represents all decision hyperplanes on the forward path from input to the activation value on the out-channel feature map; (2) We map weight vectors in the last FC layer back to input image space and visualize a single reconstructed hypersurface that decides the CNN's prediction.

As we mentioned before, AdjointBackMap requires two necessary conditions to enable it. These two necessary conditions are, (1) The CNN should not have any bias inside. Otherwise, Eq.2 would not be properly established. For example, a CNN using batch normalization [25] may not be qualified for our technique; (2) Approximation of a forward path $\boldsymbol{F}$ from input to a receptive field on an in-channel feature map should satisfy Eq.3. In detail, there exists $\boldsymbol{z}(\boldsymbol{x})$ such that the difference between the $\boldsymbol{F}(\boldsymbol{x})$ and $\boldsymbol{J_F}(\boldsymbol{z}(\boldsymbol{x}))\boldsymbol{x}$ is relatively small. In this case, care should be taken when deploying our method on a CNN using activation functions whose derivative is not piecewise constant (like tanh, proof of Section B in Appendix).

Basically, our method shows that the effective hypersurface reconstructed by AdjointBackMap is able to reproduce a CNN unit's activation given an input image through Eq.4, and the value of that activation should be the same as the one by propagating the same input through the CNN forward to the out-channel feature map or the FC layer. That is, if we dot-product this effective hypersurface to the input image directly, the returned value will precisely match the result obtained from the CNN itself. This is a crucial point that distinguishes our model from other methods developed in the field of explainable AI (table 5, 6, Appendix).

In practice, AdjointBackMap provides five reconstruction modes (RMs). Four of them act on a convolutional layer and one on the FC layer. Conv layers have four RMs due to the two factors below: (1) Whether to separate mapping along with the preset stride during training. When a kernel moves by a stride, its corresponding receptive field on the feature map will change accordingly. In short, $\boldsymbol{F}$ (Eq.3) will change when a kernel moves; (2) Whether to separate mapping along with in-channel feature maps or its corresponding in-channel kernels. This is because every in-channel feature map convolves with its corresponding in-channel kernel before merging to form an out-channel feature map.

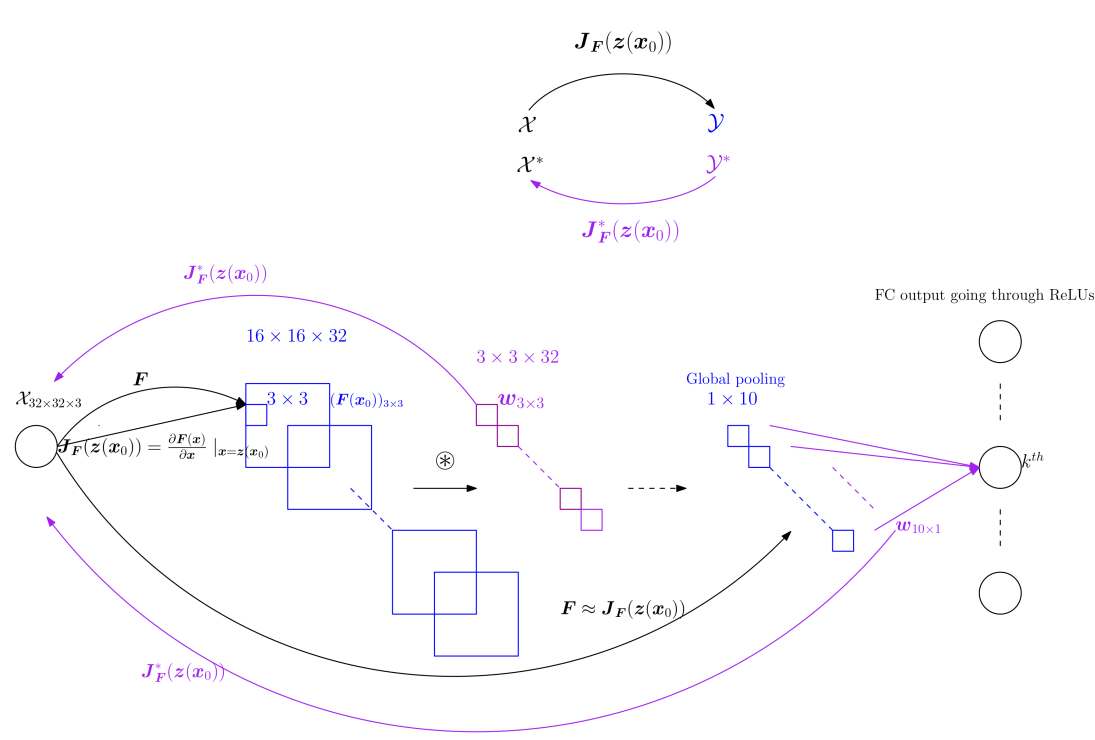

Figure 2: Figure shows geometrical relationship between $\boldsymbol{J_F}(\boldsymbol{z}(\boldsymbol{x}_0))$ and $\boldsymbol{J}^*_{\boldsymbol{F}}(\boldsymbol{z}(\boldsymbol{x}_0))$. Besides, it depicts how we apply this to either conv layer or FC layer ($\approx$ refers to Eq.3.).

We explain with an example. Suppose a CNN takes a $32 \times 32 \times 3$ (height $\times$ width $\times$ channels) RGB image $\boldsymbol{x}$. Its third convolutional layer has in-channel feature maps of shape $16 \times 16 \times 32$. That layer is equipped with convolutional kernels of $3 \times 3 \times 32 \times 64$ (height $\times$ width $\times$ in-channels $\times$ out-channels). The convolutional stride is 2 with padding 'same' [26]. We name the five RMs of AdjointBackMap as $RM_4$ to $RM_0$.

(**RM$_4$**) Convolutional kernels are separated along in-channels. Or, equivalently, each in-channel feature map is separated to convolve with its own kernel. AdjointBackMap works on kernels, one at a time ($32 \times 64$ kernels). Also, each stride move is backward mapped independently. That is every stride offset $s$ generates its own effective hypersurface of shape $32 \times 32 \times 3$, and total moves are $8 \times 8$. Therefore effective hypersurfaces reconstructed through $RM_4$ are (from Eq.5),

$$\boldsymbol{H}^{Adj}_{s,j,i}(\boldsymbol{z}(\boldsymbol{x})) = \boldsymbol{J}^*_{\boldsymbol{F}_{s,j,i}}(\boldsymbol{z}(\boldsymbol{x}))\boldsymbol{w}_{3\times3,s,j,i}, \tag{6}$$

where $s \in \{0, 1, ..., 63\}, j \in \{0, 1, ..., 31\}, i \in \{0, 1, ..., 63\}$, $\boldsymbol{F}_{s,j,i}$ denotes a forward path from the input to the receptive field offset by stride $s$ on the $(j, i)^{th}$ kernel. In this case, the number of backward mappings is $8 \times 8 \times 32 \times 64$, and each one has its shape of $32 \times 32 \times 3$. This is the most basic application that strictly follows our theory. We are able to visualize what a single kernel is actually doing, from the perspective of input image space, on its effective receptive field with a stride shift.

(**RM$_3$**) Convolutional kernels are separated along in-channels. However, each stride sums together to form a single backward mapping. The total kernels are $32 \times 64$ and each kernel has its shape of $3 \times 3$. Each stride reconstructs its own effective hypersurface of shape $32 \times 32 \times 3$ and total moves are $8 \times 8$. Then all $8 \times 8$ moves sums together pixel-wise to form a single $32 \times 32 \times 3$ effective hypersurface. This is an unusual scope (usual scope for the layer right before global pooling [27]), which means we get an effective hypersurface $\boldsymbol{H}_{j,i}^{Adj}(\boldsymbol{z}(\boldsymbol{x}))$ in this way,

$$\langle \boldsymbol{x}, \boldsymbol{H}_{j,i}^{Adj}(\boldsymbol{z}(\boldsymbol{x}))\rangle = \sum_{s=0}^{8\times 8-1} (\boldsymbol{F}(\boldsymbol{x}) \circledast \boldsymbol{w}_{3\times 3,s,j,i}) = \langle \boldsymbol{x}, \sum_{s=0}^{63} \boldsymbol{H}_{s,j,i}^{Adj}(\boldsymbol{z}(\boldsymbol{x}))\rangle. \tag{7}$$

The relationship to $RM_4$ is supported by Eq.5 and linearity in duality and is pointed out by the last "=". In this case, the number of backward mappings is $32 \times 64$, and each one has its shape of $32 \times 32 \times 3$.

($\mathbf{RM_2}$) Convolutional kernels are not separated along in-channels anymore, but each stride is still backward mapped independently. Thus the number of backward mappings is $8 \times 8 \times 64$, and each one has its shape of $32 \times 32 \times 3$. Effective hypersurfaces reconstructed from $RM_2$ are,

$$\begin{aligned} \boldsymbol{H}_{s,i}^{Adj}(\boldsymbol{z}(\boldsymbol{x})) &= \sum_{j=0}^{31} \boldsymbol{J}^*_{\boldsymbol{F}_{s,j,i}}(\boldsymbol{z}(\boldsymbol{x}))\boldsymbol{w}_{3\times 3,s,j,i} \\ &= \sum_{j=0}^{31} \boldsymbol{H}_{s,j,i}^{Adj}(\boldsymbol{z}(\boldsymbol{x})). \end{aligned} \tag{8}$$

Similarly, the last "=" reveals its relationship to $RM_4$. This effective hypersurface decides a single unit's activation in the out-channel feature map.

($\mathbf{RM_1}$) Convolutional kernels are not separated along in-channels. Each stride move also sums together to form a single backward mapping as well. Thus the total quantity of backward mappings is 64, and each one has its shape of $32 \times 32 \times 3$. Similar to $RM_3$, this is an unusual scope (usual for the layer right before global pooling), which means we get an effective hypersurface $\boldsymbol{H}_{i}^{Adj}(\boldsymbol{z}(\boldsymbol{x}))$ in this way,

$$\begin{aligned} \langle \boldsymbol{x}, \boldsymbol{H}_{i}^{Adj}(\boldsymbol{z}(\boldsymbol{x}))\rangle &= \sum_{s=0}^{63}(\boldsymbol{F}_s(\boldsymbol{x}) \circledast \boldsymbol{w}_{s,3\times 3\times 32\times 64}) \\ &= \langle \boldsymbol{x}, \sum_{j=0}^{31} \boldsymbol{H}_{j,i}^{Adj}(\boldsymbol{z}(\boldsymbol{x}))\rangle = \langle \boldsymbol{x}, \sum_{s=0}^{63} \boldsymbol{H}_{s,i}^{Adj}(\boldsymbol{z}(\boldsymbol{x}))\rangle. \end{aligned} \tag{9}$$

In other words, the inner product between an input image and the effective hypersurface is equal to the pixel-wise summation of the out-channel feature map. Relations to $RM_3$ and $RM_2$ are listed in the second line of Eq.9.

**($\mathbf{RM_0}$)** The backward mapping deploys on the weight vectors $\{\boldsymbol{w}_k\}$ where $k$ denotes class index in the FC layer of the CNN. Then the output value for class $k$ (before going through an activation) is decided by an effective hypersurface $\boldsymbol{H}_{k}^{Adj}(\boldsymbol{z}(\boldsymbol{x}))$ of shape $32 \times 32 \times 3$, i.e.,

$$\boldsymbol{H}_{k}^{Adj}(\boldsymbol{z}(\boldsymbol{x})) = \boldsymbol{J}^*_{\boldsymbol{F}}(\boldsymbol{z}(\boldsymbol{x}))\boldsymbol{w}_k, \tag{10}$$

where $k \in \{0, 1, ..., (N-1)\}$, $N$ denotes the number of classes.

**Implementation:** We use convolution to compute duality in Eq.5 as they are equivalent, and duality can use hardware acceleration. Due to computationally expensive Jacobian, Eq.6 $\sim$ Eq.10 are optimized and summarized in Algorithm 1. Note conv2d, unstack, stack, expanddim, matmul are functions defined in Tensorflow [26]. Padding of conv2 is the same as training. Also, conv2d has an 'axis' choice in order to replace the transpose in Eq.5.

Though an effective hyperplane acquired from AdjointBackMap is in the same space as the original input image, the scalar element values of the effective hyperplane might not lie on the same interval as its original image, which has values in $[0, 1]$. In that case, we have to properly normalize the value to enable its visualization. We will explain more in our experiments.

# 5 Experiments

## 5.1 Pre-trained Model

We discuss below how we train our CNN models.

**Dataset:** We used CIFAR10 [28] as our dataset. CIFAR10 contains 50k $32 \times 32$ RGB (value range $[0, 1]$) images for training and 10k for testing, and classes are 10. All images are normalized with RGB means: $0.4914, 0.4822, 0.4465$, and standard deviations: $0.2023, 0.1994, 0.2010$ [29]. We separated the 50k samples into one training set and one validation set with a ratio of $9:1$, i.e. 45k, 5k, respectively. All analyses were conducted on the test set.

**Algorithm 1** AdjointBackMap from $RM_4$ to $RM_0$

**Input:**

1. $\boldsymbol{x}_d$: input ($d = H \times W \times C$);
2. $\boldsymbol{z}$: function for Eq. 3;
3. $\boldsymbol{T}$: pre-trained model;
4. $i$: layer index;
5. $s$: stride during training;
6. $L$: RM number.

**Output:** Effective hypersurface $\boldsymbol{H}^{Adj}(\boldsymbol{z}(\boldsymbol{x}_d))$

```
function AdjointBackMap(x_d, z, T, i, s, L)
    z_0 = z(x_d)
    switch L do

    case 'RM_0':
    load w_{c_in × c_labels} from FC layer of T
    load F_{c_in} from T
    J_F = ∂F_{c_in} / ∂x_d
    return matmul(J_F(z_0), w_{c_in × c_labels}, axis=c_in)
    end case

    case 'RM_4' or 'RM_3':
    load w_{r_1 × r_2 × c_in × c_out} from T at i
    load F_{H_i × W_i × c_in} from T at i
    J_{F,d×H_i×W_i×c_in} = ∂F_{H_i×W_i×c_in} / ∂x_d
    w_u = unstack(w_{r_1 × r_2 × c_in × c_out}, axis=c_in)
    J_{F,u} = unstack(J_{F,d×H_i×W_i×c_in}, axis=c_in)
    Empty container R, j = 0
    while j < c_in do
        J_F = expanddim(J_{F,u}[j], axis=c_in)
        w = expanddim(w_u[j], axis=c_in)
        R.append(conv2d(J_F(z_0), w, stride=s, axis=(H_i, W_i, c_in, c_out)));
        j = j + 1
    end while
    conv_{d×⌈H_i/s⌉×⌈W_i/s⌉×c_in×c_out} = stack(R, axis=c_in)
    if L is 'RM_4' then
        return conv_{d×⌈H_i/s⌉×⌈W_i/s⌉×c_in×c_out}
    else if L is 'RM_3' then
        return sum(conv_{d×⌈H_i/s⌉×⌈W_i/s⌉×c_in×c_out}, axis=(⌈H_i/s⌉, ⌈W_i/s⌉))
    end if
    end case

    case 'RM_2' or 'RM_1':
    load w_{r_1 × r_2 × c_in × c_out} from T at i
    load F_{H_i × W_i × c_in} from T at i
    J_F = ∂F_{H_i×W_i×c_in} / ∂x_d
    conv_{d×⌈H_i/s⌉×⌈W_i/s⌉×c_out} = conv2d(J_F(z_0),
      w_{r_1 × r_2 × c_in × c_out}, stride=s, axis=(H_i, W_i, c_in, c_out))
    if L = 'RM_2' then
        return conv_{d×⌈H_i/s⌉×⌈W_i/s⌉×c_out}
    else if L = 'RM_1' then
        return sum(conv_{d×⌈H_i/s⌉×⌈W_i/s⌉×c_out}, axis=(⌈H_i/s⌉, ⌈W_i/s⌉))
    end if
    end case
    end switch
end function
```

**Data Augmentation:** We used data augmentation for training. An input color image goes through randomly flipping of left to right, random adjustment of saturation within $[0.0, 2.0]$, random adjustment of contrast within $[0.4, 1.6]$, random adjustment of brightness with 0.5, resizing to $36 \times 36 \times 3$ and then randomly cropped to $32 \times 32 \times 3$.

**Model:** We used two models: (1) VGG [30] with 7 activation layers (VGG7, without bias) and (2) 20-layer ResNet with fixup initialization [29] [31](Fixup-ResNet20, with weights rescaling and without bias, the same as Figure 1 (Middle) of [29]). The learnable parameters of VGG7 and Fixup-ResNet20 are listed in table 1 and 3 (Appendix), respectively.

**Cost and Accuracy:** We regularized the kernels by $L_1$ (factor $10^{-4}$). We used cross-entropy with softmax as a cost. Accuracy was measured by a prediction index being exactly matched with its label (Top-1 accuracy).

**Training, Validation, Test:** We trained and validated VGG7 and Fixup-ResNet20 with GD (gradient descent) optimizer on an RTX2080Ti. The batch size was 100. We trained VGG7 for a total of 301 epochs. We set the learning rate to $2 \times 10^{-4}$ at the start of training and dropped it to $10^{-4}$, $5 \times 10^{-5}$ at the $200^{th}$, the $250^{th}$ epoch, respectively. We trained Fixup-ResNet20 for a total of 201 epochs. We set the learning rate to $2 \times 10^{-3}$ at the start of training and dropped it to $10^{-3}$, $5 \times 10^{-4}$ at the $100^{th}$, the $150^{th}$ epoch, respectively. We trained with different learning rates or epochs so that both models could learn sufficiently. We trained with 45k samples every epoch and validated the trained model on 5k samples every two epochs, and the trained model will be saved if a higher validation accuracy is detected. VGG7 and Fixup-ResNet20 reports test accuracy of 85.6% and 90.3%, respectively (10k test samples).

## 5.2 Five Experiments With Respect to Five RMs

We first verify the existence of a function $\boldsymbol{z}$ in Eq.3 (figure 7 and 8 in Appendix report our AdjointBackMap precisely reconstructs the effective hypersurfaces with over 99.99% of the relative errors $\leq 0.01$ in both convolutional layers and FC layers.). Then we move to five experiments related to five RMs ($RM_4$ to $RM_1$ are shown in the Appendix). We built TensorFlow 1.15.4 from the source code and enabled its functionality on AVX-2, AVX-512, FMA3 instruction sets to speed up all experiments. In Appendix, figure 7 and 8 computation respectively took 30 and 335 hours on a 9940X CPU with 128 gigabytes (GBs) DRAM; Simulation of either figure 10 or 12 roughly consumed 200 GBs DRAM on a 10920X CPU.

### 5.2.1 Visualization of $\mathbf{RM_0}$:

$RM_0$ works on weights of FC layer to generate a set $\{\boldsymbol{H}_k^{Adj}(\boldsymbol{z}(\boldsymbol{x}_0)) \mid k \in \{0, 1, ..., 9\}\}$ (Eq.10). Each $\boldsymbol{H}_k^{Adj}(\boldsymbol{z}(\boldsymbol{x}_0))$ represents an effective hyperplane that decides the $k^{th}$ value of the FC output (before ReLU6). That is, the VGG7 (or Fixup-ResNet20) taking an image and passing forward through layers to make a prediction is equivalent to doing ten inner products between the image and our $\{\boldsymbol{H}_k^{Adj}(\boldsymbol{z}(\boldsymbol{x}_0)) \mid k = \{0, ..., 9\}\}$. $\boldsymbol{H}_k^{Adj}(\boldsymbol{z}(\boldsymbol{x}_0))$ has its shape of $32 \times 32 \times 3$. We apply the same visualization techniques as $RM_4$ (Appendix).

**Results:** We use those two images (ship, frog) from $RM_4$ (Appendix). Also, we add one image that has the same ship label. The results are shown in Figure 3.

# 6 Analysis of Experiments

As the layer goes high, the effective receptive field (non-black pixels or non-zero area) enlarges. This is obvious from the plots of both $RM_4$ and $RM_2$ (Appendix). The reason is if a kernel in Conv0 has a shape of $3 \times 3$, the same size kernel on Conv1 actually will have a maximal effective receptive field of $5 \times 5$ from the original input perspective because of stride offset $s = 1$. However, few kernels do fully utilize their effective receptive fields. Even, some layer has a kernel that takes null from the input image (its effective receptive field blacks out) and therefore makes $\boldsymbol{H}_{\{s\},j,i}^{Adj}(\boldsymbol{z}(\boldsymbol{x}_0))$ (eq.20, Appendix) sparse.

Effective hyperplanes from $RM_4$ to $RM_0$ are not human recognizable patterns. For example, patterns of $H_{k=8}^{Adj}(\boldsymbol{z}(\boldsymbol{x}_0))$ in Figure 3(a, d, c, f) show neither a clear ship nor a rough contour of a ship. Patterns of Figure 3(a) (or c) are significantly different from Figure 3(d) (or f), although they are the same class (ship). $RM_3$ and $RM_1$ (Appendix) have coherent colored shapes at low-level convolutional layers; However, they gradually turn to irregular pixels as the layer goes high. Even the same kernel shows different AdjointBackMap patterns as the stride moves. That means a kernel may make a different decision from the input image's perspective when it moves to a different receptive field though the kernel itself does not physically change at all during the stride move. These imply CNN's decision is sensitive to values in each pixel of the input image.

As the proof (Section B, Appendix) of Eq.3 reveals, changing an input image $\boldsymbol{x}$ forces $\boldsymbol{H}^{Adj}(\boldsymbol{z}(\boldsymbol{x}))$ to vary at the same time. This has been experimentally tested in figure 7, 8 (Appendix). Specifically, an input image $\boldsymbol{x}$ goes through VGG7 or Fixup-ResNet20 $\boldsymbol{F}$ to get an output value from the FC layer, $\boldsymbol{F}_k(\boldsymbol{x})$, for the $k^{th}$ class, will have,

$$\boldsymbol{F}_k(\boldsymbol{x}) = \langle \boldsymbol{x} \mid \boldsymbol{h}_k(\boldsymbol{x}) \rangle, \tag{11}$$

where $\boldsymbol{h}_k(\boldsymbol{x})$ is the effective hypersurface for the $k^{th}$ output from FC layer. Therefore, the CNN decision is largely conditioned on input.

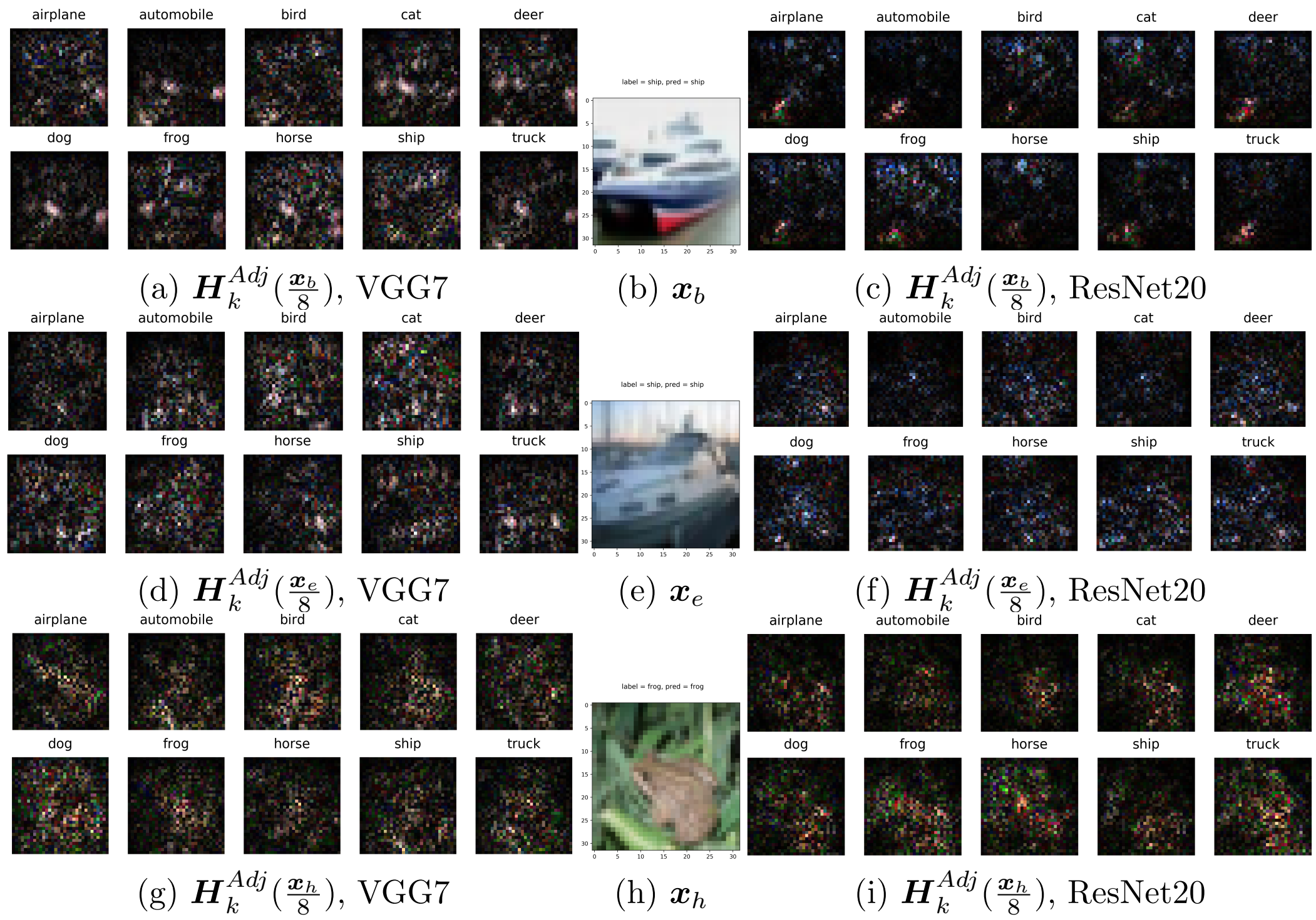

(a) $\boldsymbol{H}_k^{Adj}(\frac{\boldsymbol{x}_b}{8})$, VGG7 (b) $\boldsymbol{x}_b$ (c) $\boldsymbol{H}_k^{Adj}(\frac{\boldsymbol{x}_b}{8})$, ResNet20

(d) $\boldsymbol{H}_k^{Adj}(\frac{\boldsymbol{x}_e}{8})$, VGG7 (e) $\boldsymbol{x}_e$ (f) $\boldsymbol{H}_k^{Adj}(\frac{\boldsymbol{x}_e}{8})$, ResNet20

(g) $\boldsymbol{H}_k^{Adj}(\frac{\boldsymbol{x}_h}{8})$, VGG7 (h) $\boldsymbol{x}_h$ (i) $\boldsymbol{H}_k^{Adj}(\frac{\boldsymbol{x}_h}{8})$, ResNet20

Figure 3: Visualization of $\mathbf{RM_0}$ on VGG7 and Fixup-ResNet20. $\boldsymbol{H}_k^{Adj}(\frac{\boldsymbol{x}_0}{8})$ patterns mapped from the FC layer (Eq.10) by $RM_0$ (Six figures. VGG7: a, d, g; Fixup-ResNet20: c, f, i.), and their corresponding inputs (b, e, h, respectively). The number of subfigures in a plot is equal to the number of classes.

# 7 Applications to Adversarial Inputs

A well-known adversarial attack from [9] has been discussed in Section 1. In this section, we apply AdjointBackMap to analyze adversarial inputs and further probe the functional properties of CNNs through three experiments: $\mathbf{A}, \mathbf{B_1}, \mathbf{B_2}$. We use $\boldsymbol{Advr}$ to denote an adversarial "noise".

**Experiment A:** We visualize the fluctuations of effective hypersurfaces (in a CNN model) under an untargeted attack. We use the same basic iterative method as [32] [9] to compute an adversarial noise pattern. The factor in [9] is 0.007. We use a factor of 0.04 (around 5×) because we normalize an input image with the stds around 0.2. Computed "noise" is added to the input image and fools VGG7 from "horse" to "dog" (We repeat the experiment for Fixup-ResNet20.). We first have to verify eq.18 (Appendix) before our visualization: results (table 5 and 6, Appendix) show the $M_3$ column has bigger approximation errors than $M_2$. Even, the $M_3$ in table 6 incorrectly predicts the class. Thus, we confirm that only $\{\boldsymbol{H}_k^{Adj}(\frac{\boldsymbol{x}_0+\boldsymbol{Advr}}{8}) \mid k \in \{0,1,...,9\}\}$ is the set of effective hyperplanes that decide the CNN predictions on $(\boldsymbol{x}_0 + \boldsymbol{Advr})$ for 10 classes instead of $\{\boldsymbol{H}_k^{Adj}(\frac{\boldsymbol{x}_0}{8}) \mid k \in \{0,1,...,9\}\}$. Then, we visualize the differences of effective hyperplanes between a normal input and its adversarial one.

**Experiment $\mathbf{B_1}$:** We explore the fluctuations of a single hypersurface under targeted attacks. We use the iterative least-like method as [32] and make 9 adversarial noises for the horse image, $\{\boldsymbol{Advr}_i \mid i \in \{0,1,..,9\}, \boldsymbol{Advr}_7 = \theta\}$ (name $S_{B_1}$, where 7 is the label index), to mislead our VGG7 to 9 other classes. We project $\{\boldsymbol{H}_{k=7}^{Adj}(\frac{\boldsymbol{x}_0+\boldsymbol{Advr}_i}{8}) \mid i \in \{0,1,...,9\}\}$ with tSNE [33] for analysis. We repeat for Fixup-ResNet20 as well.

**Experiment $\mathbf{B_2}$:** Further, we show an effective hypersurface under down-scaled adversarial noises is different from the one under Gaussian noise. We generate a set, $S_{B_2} = \{\beta_j \times \boldsymbol{advr} | j < 50, j \in \mathbb{N}, \boldsymbol{advr} \in S_{B_1}\}\} \cup \{g_m \mid m < 50, m \in \mathbb{N}; g_m \sim \mathcal{N}(\mu = 7.147 \times 10^{-4}, \sigma^2 = 2.487 \times 10^{-5})\}$ , containing 100 "noisy" samples for VGG7, using the following procedures. First, we sequentially iterate $\boldsymbol{advr}_i \in S_{B_1}$ for 9 epochs (skip $i = 7$); Each epoch, if two conditions:

1. Misclassified: predicted class (VGG7) on $(\boldsymbol{x}_0 + \beta \times \boldsymbol{advr}_i)$ is not equal to the label of $\boldsymbol{x}_0$,

2. Overthreshold: predicted value from the class is greater than (0.5 (threshold) + predicted value) for the label,

are satisfied in looping $\beta$ from 1.0 to 0 with a step=$-0.05$, a qualified $\beta \times \boldsymbol{advr}_i$ will be stored; We randomly shuffle before filtering 75 qualified ones to 50. Then, we generate 50 Gaussian noise ($32 \times 32 \times 3$) using the same statistical pixel mean and variance computed from $S_{B_1}$. And none of the Gaussian, when applied to the original image, will fool the VGG7 prediction. We unite them together as $S_{B_2}$ and project $\{\boldsymbol{H}_{k=7}^{Adj}(\frac{\boldsymbol{x}_0+n}{8}) \mid n \in S_{B_2}\}$ with Factor Analysis [33]. We repeat the experiment for Fixup-ResNet20 with $\mu = 2.892 \times 10^{-6}$, $\sigma^2 = 9.861 \times 10^{-6}$ and 68 qualified ones to filter.

**Results:** Results of Experiment **A** are illustrated in Figure 4; Results of Experiment $\mathbf{B_1}$, $\mathbf{B_2}$ are illustrated in Figure 5, 6, respectively. Details below.

**Analysis:** Experiment **A** reveals that values of $\boldsymbol{H}_k^{Adj}(\frac{\boldsymbol{x}_0+\boldsymbol{Advr}}{8})$ are significantly different from $\boldsymbol{H}_k^{Adj}(\frac{\boldsymbol{x}_0}{8})$ (Figure 4($a_3$ or $b_3$)). We learn from ($a_4$ or $b_4$) that this difference starts from Conv1, the second convolutional layer, through our $RM_3$, though no kernel in that layer changes. It implies effective hypersurfaces are very sensitive in response to input and either VGG7 or Fixup-ResNet20 takes different roads to decide on two visually identical images, which is different from humans who may ignore small variation in pixel value.

As we have seen in Figure 3, despite the same class, (a) and (d) (or (c) and (f)) have different patterns. Then Experiment $\mathbf{B_1}$ (Figure 5) reveals that a single effective hypersurface is brittle to different adversarial noises despite being visually indistinguishable. Further, Experiment $\mathbf{B_2}$ (Figure 6) discloses the effective hypersurface is still weak under a set of linearly scaled adversarial input. These imply that visually similar images can easily knock off a CNN by misleading its decision process, which takes advantage of the fact that CNN is essentially different from our human vision.

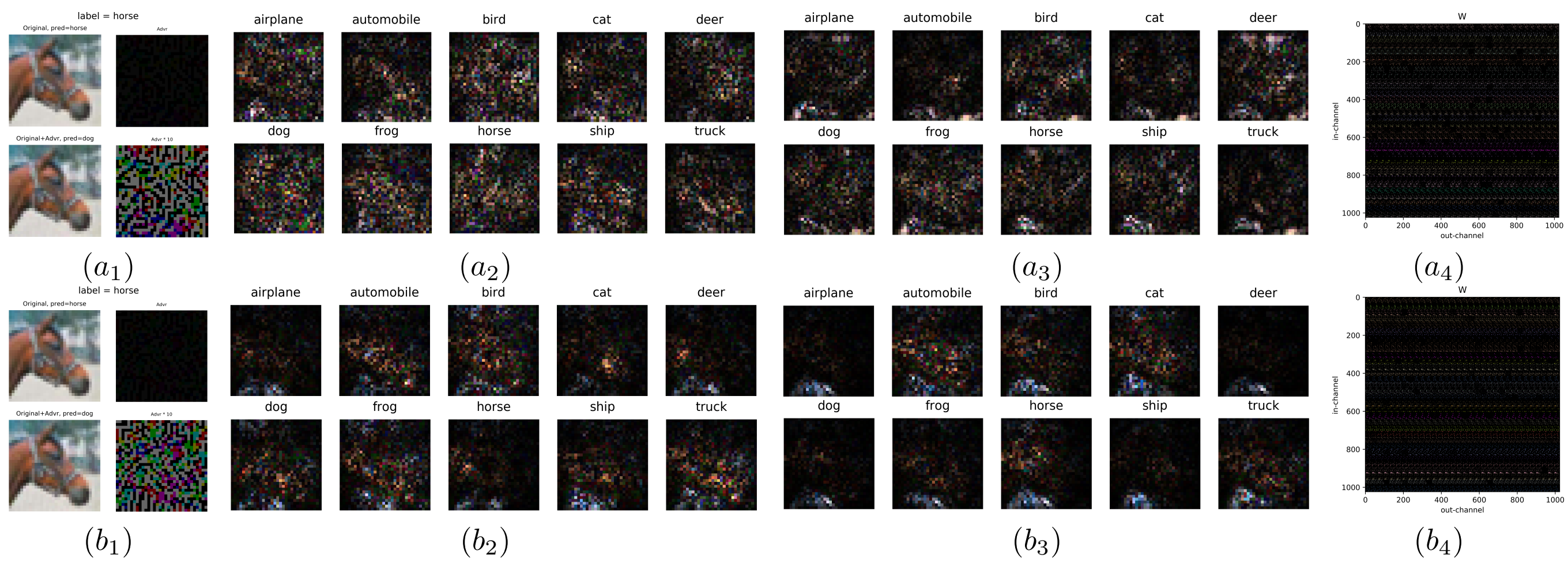

Figure 4: Experiment **A**: Visualize misclassification of VGG7 on its adversarial example ("horse" to "dog"). $\boldsymbol{z}_0 = \frac{\boldsymbol{x}_0+\boldsymbol{Advr}}{8}$, $\boldsymbol{z}_1 = \frac{\boldsymbol{x}_0}{8}$. ($a_1$) Original ($\boldsymbol{x}_0$) + $\boldsymbol{Advr}$; ($a_2$) $\{\boldsymbol{H}_k^{Adj}(\boldsymbol{z}_0) \mid k \in \{0,1,...,9\}\}$ by $RM_0$; ($a_3$) $\{\boldsymbol{H}_k^{Adj}(\boldsymbol{z}_0) - \boldsymbol{H}_k^{Adj}(\boldsymbol{z}_1)\}$ by $RM_0$; ($a_4$) $(\boldsymbol{H}_{\{j,i\}}^{Adj}(\boldsymbol{z}_0) - \boldsymbol{H}_{\{j,i\}}^{Adj}(\boldsymbol{z}_1))$ by $RM_3$ on Conv1. The experiment is repeated on Fixup-ResNet20 and listed as $b_1 \sim b_4$ ($b_4$ by $RM_3$ on Conv1 of Residual Block 0). ($a_3$) and ($a_4$) (or $b_3$ and $b_4$) are comparisons (between $\boldsymbol{z}_0$ and $\boldsymbol{z}_1$). ($a_4$) (or $b_4$) implies that two visually identical images have different effective hyperplanes starting from the Conv1 layer, although their kernels in that layer are the same. Therefore, effective hypersurfaces are very sensitive to input.

# 8 Conclusions & Future Work

We introduce adjoint-operator-based AdjointBackMap, which maps a kernel or a weight vector back to the dual of input space as an effective decision boundary. Using Riesz representation, we are able to project them back to the input space to enable visualization. AdjointBackMap works as long as certain conditions are satisfied. Through five reconstruction modes of AdjointBackMap, we visualize the effective decision hyperplanes at different layers in the CNN and find they are different from human vision. Also, we learn that a CNN's decisions are sensitive to small changes in the input. We expect our work to motivate principled approaches to explainable AI and adversarial attacks.

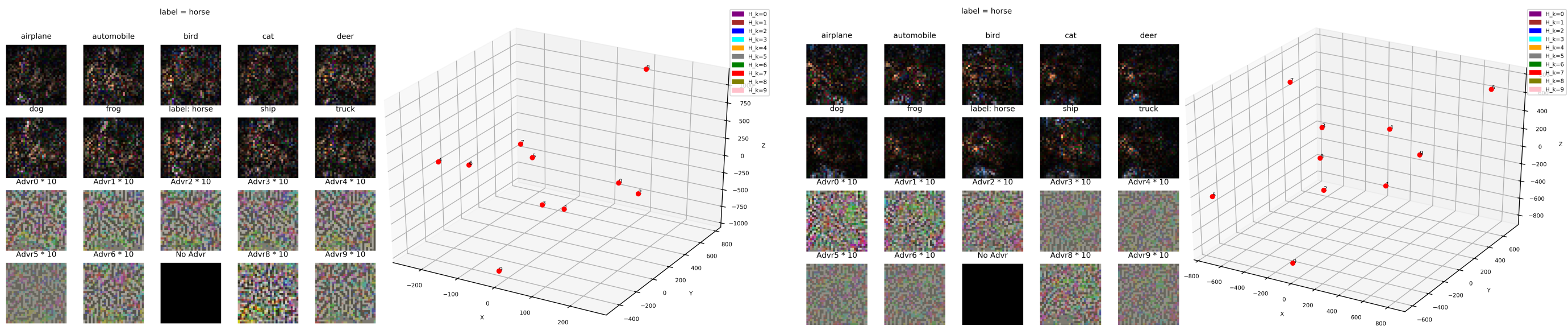

$(a_1)$ $(b_1)$ $(a_2)$ $(b_2)$

Figure 5: Experiment $\mathbf{B_1}$: Visualize an effective hypersurface under target attack and its tSNE analysis. $(a_1)$ $\{\boldsymbol{H}_{k=7}^{Adj}(\frac{\boldsymbol{x}_0+\boldsymbol{Advr}_i}{8}) \mid \boldsymbol{Advr}_i \in S_{B_1}\}$ by $RM_0$; $(b_1)$ tSNE reducing $\{\boldsymbol{H}_{k=7}^{Adj}(\frac{\boldsymbol{x}_0+\boldsymbol{Advr}_i}{8}) \mid \boldsymbol{Advr}_i \in S_{B_1}\}$. We prepare 9 adversarial noise, $S_{B_1} = \{\boldsymbol{Advr}_i \mid i \in \{0,1,..,9\}, \boldsymbol{Advr_7} = \theta\}$, for the horse image (Figure 4$(a_1)$). An element $\boldsymbol{Advr}_i$ can fool VGG7 to the class index $i$. $\boldsymbol{Advr_7} = \theta$ denotes no (zero) adversarial noise. Patterns of $\boldsymbol{H}_{k=7}^{Adj}(\frac{\boldsymbol{x}_0+\boldsymbol{Advr}_i}{8})$ are sequentially subfigured in first two rows of $(a_1)$ and rest two rows show its corresponding $\boldsymbol{Advr}_i$ magnified by 10. These 9 visually indistinguishable adversarial noise have small euclidean distances, $\{\|\boldsymbol{Advr}_i\|_2\} = \{3.822, 3.821, 3.839, 3.809, 3.913, 2.217, 3.821, 0, 5.465, 3.838\}$, compared with $\|\boldsymbol{x}_0\| = 58.856$. We project the metric relationship among 10 effective hyperplanes with tSNE $(b_1)$. The experiment is repeated on Fixup-ResNet20 $(a_2, b_2)$ with $\{\|\boldsymbol{Advr}_i\|_2\} = \{4.685, 4.09, 3.381, 2.217, 2.217, 2.217, 2.217, 0, 3.484, 2.217\}$. Visualization has illustrated that differences exist in the effective hyperplanes ((a) and (d) for VGG7 or (c) and (f) for Fixup-ResNet20 in Figure 3) for two images from the same class. Now $(a_1)$ and $(b_1)$ (or $(a_2)$ and $(b_2)$) show that effective hyperplanes for two indistinguishable images are still far apart from each other. Thus, CNN's decision is brittle.

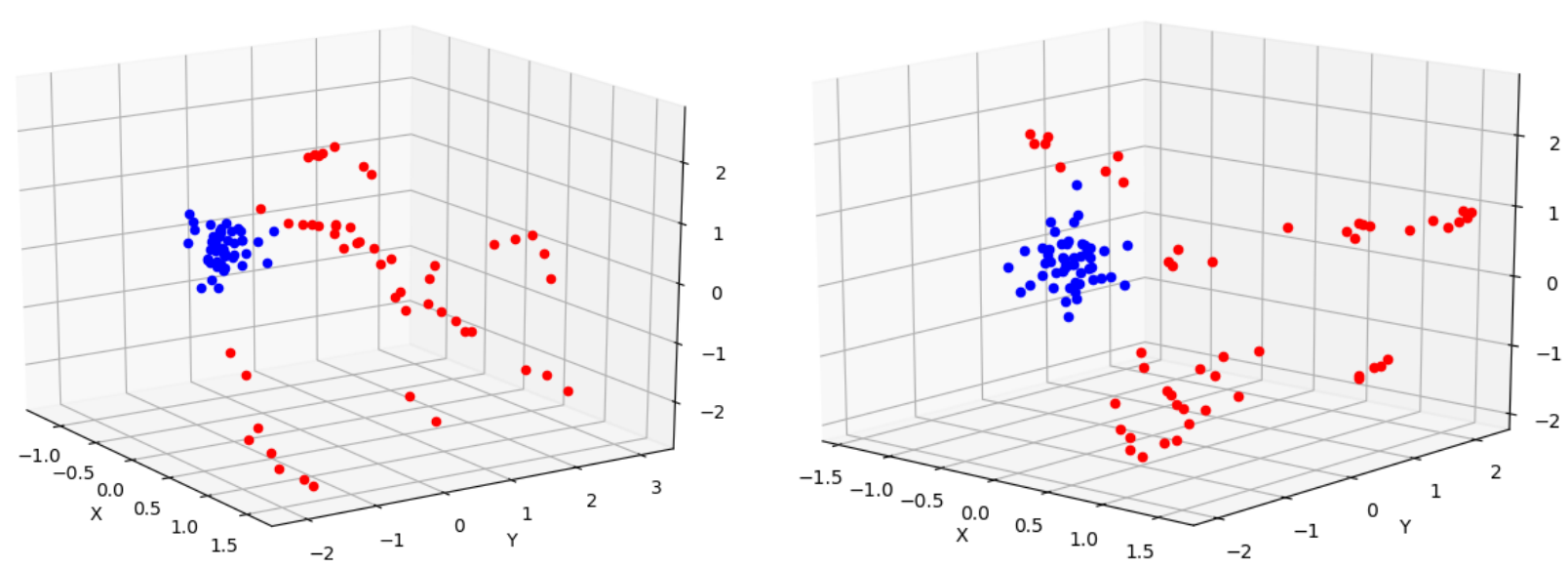

(a) VGG7 (b) Fixup-ResNet20

Figure 6: Experiment $\mathbf{B_2}$: Factor Analysis of the effective hypersurface with down-scaled adversarial noise in contrast to Gaussian noise. $S_{B_2}$ computed from 50 adversarial examples (from $S_{B_1}$, red) and 50 Gaussian noise (same mean and variance as $S_{B_1}$) added non-adversarial samples (blue). Other than Experiment $\mathbf{B_1}$ (Figure 5), we analyze causality between misclassified adversarial examples and correctly classified Gaussian noises). Effective hypersurfaces of adversarial cases (red) are more spread out than that of non-adversarial ones (blue).

# Appendix A Notations

We will use the following notations.

1. "Eq" refers to an equation in the main paper;
2. "eq" refers to an equation in the appendix;
3. "Algorithm" refers to an algorithm in the main paper;
4. "table" refers to a table in the appendix;
5. "Figure" refers to a figure in the main paper;
6. "figure" refers to a figure in the appendix;
7. RM(s) is an abbreviation of Reconstruction Mode(s) from the main paper.

# Appendix B Proof of Eq.3

We prove Eq.3 holds for $\boldsymbol{z}(\boldsymbol{x}) = k\boldsymbol{x}, k \in \mathbb{R}^+$ when the neural network (without any bias) is activated with either ReLU or Leaky ReLU. Since the rest are trivial, we only prove $\boldsymbol{F}(\boldsymbol{x}) = \boldsymbol{J}_{\boldsymbol{F}}(\boldsymbol{z}(\boldsymbol{x}))\boldsymbol{x}$.

For any single convolutional layer $l$, without loss of generality, a pixel $p$ of the feature map is activated after in-channel kernels $\boldsymbol{w}_{r_1 \times r_2, l}$ convolving on its in-channel receptive field feature maps $\boldsymbol{x}_{r_1 \times r_2, l-1}$ (Eq.4, 5),

$$\begin{aligned} c &= \sum_j \langle \boldsymbol{x}_{j, r_1 \times r_2, l-1} \mid \boldsymbol{w}_{j, r_1 \times r_2, l} \rangle, \\ p &= \sigma(c), \end{aligned} \tag{12}$$

where $j$ denotes in-channel index and $\sigma$ is an activation of either ReLU or Leaky ReLU. Also,

$$\sum_j \langle \boldsymbol{x}_{j, r_1 \times r_2, l-1} \mid \frac{\partial c}{\partial \boldsymbol{x}_{j, r_1 \times r_2, l-1}} \rangle = \sum_j \langle \boldsymbol{x}_{j, r_1 \times r_2, l-1} \mid \boldsymbol{w}_{j, r_1 \times r_2, l} \rangle = c. \tag{13}$$

We have,

$$\begin{aligned} &\sum_j \langle \boldsymbol{x}_{j, r_1 \times r_2, l-1} \mid \frac{\partial p}{\boldsymbol{x}_{j, r_1 \times r_2, l-1}} \Big|_{\boldsymbol{z}(\boldsymbol{x}_{r_1 \times r_2, l-1})} \rangle \\ &= \sum_j \langle \boldsymbol{x}_{j, r_1 \times r_2, l-1} \mid \frac{\partial p}{\partial c} \boldsymbol{w}_{j, r_1 \times r_2, l} \rangle \\ &= \begin{cases} r, & kc < 0 \\ c, & kc \geq 0 \end{cases}, k \in \mathbb{R}^+ \\ &= \sigma(c) = p. \end{aligned} \tag{14}$$

where $r = 0$ for the ReLU or $r = -0.2 \times c$ for the default Leaky ReLU in TensorFlow. Hence, $\boldsymbol{F}_l(\boldsymbol{x}_{r_1 \times r_2, l-1}) = \boldsymbol{J}_{\boldsymbol{F}_l}(\boldsymbol{z}(\boldsymbol{x}_{r_1 \times r_2, l-1}))\boldsymbol{x}_{r_1 \times r_2, l-1}$ holds for any receptive field of any convolutional layer. Also it holds for any avg or max-pooling layer (An avg-pool is equivalent to apply a convolutionally averaging kernel). Then stacked layers of convolutions are equivalent to cascade multiplications of jacobian matrice, which ensures $\boldsymbol{F}(\boldsymbol{x}) = \boldsymbol{J}_{\boldsymbol{F}}(\boldsymbol{z}(\boldsymbol{x}))\boldsymbol{x}$ point wisely works for any input image $\boldsymbol{x}$.

We mention two things:

1. The proof also works for those activations having piecewise constant derivatives when $k = 1$ is applied;
2. Although our theory requires no bias assumption, it does not imply we cannot have any bias at all. Alternatively, we can alleviate the limitation with equivalent "bias" through input. For example, a CNN $F$ takes an affine image $\boldsymbol{x}^{'}$ (the normalization of an image $\boldsymbol{x}$),
$$\boldsymbol{x}^{'} = k\boldsymbol{x} + b, k > 0, \tag{15}$$
and predicts through two output neurons to classify binary cases $\{0, 1\}$. By the above proof and Eq.11, its decision boundary $\mathcal{C}$ is,
$$\begin{aligned} \mathcal{C} &= \{\boldsymbol{x} \mid \langle k\boldsymbol{x} + b \mid h_{k=0}(\boldsymbol{x}^{'}) \rangle \geq \langle k\boldsymbol{x} + b \mid h_{k=1}(\boldsymbol{x}^{'}) \rangle\} \\ &= \{\boldsymbol{x} \mid k\langle \boldsymbol{x} \mid h_{k=0}(\boldsymbol{x}^{'}) - h_{k=1}(\boldsymbol{x}^{'}) \rangle + b \times (h_{k=0}(\boldsymbol{x}^{'}) - h_{k=1}(\boldsymbol{x}^{'})) \geq 0\}. \end{aligned} \tag{16}$$
Here $b \times (h_{k=0}(\boldsymbol{x}^{'}) - h_{k=1}(\boldsymbol{x}^{'}))$ serves as an equivalent bias for the CNN's decision boundary. And this equivalent form allows some bias to flow in our theory.

# Appendix C Five Experiments With Respect to Five RMs

As mentioned in the main paper, we verify the existence of a function $\boldsymbol{z}$ in Eq.3. Then we move to the four remaining experiments related to the four RMs ($RM_4$, $RM_3$, $RM_2$, $RM_1$).

## C.1 Verify The Existence of $\boldsymbol{z}$

Our AdjointBackMap needs two necessary conditions. The first one has been satisfied (no bias condition). We just verify the second one.

Generally, we use Eq.3 and 4 to verify the second condition. We verify two types of tasks:

1. Verify the existence of $\boldsymbol{z}$ for the four RMs on every convolutional layer;
2. Verify the existence of $\boldsymbol{z}$ for the $RM_0$ on the FC layer.

**Verify the existence of $\boldsymbol{z}$ on every conv layer:** A convolutional layer (except the first layer) is equipped with kernels, $\boldsymbol{w}_{r_1 \times r_2 \times c_{in} \times c_{out}}$, which convolves the in-channel feature maps $(\boldsymbol{F}(\boldsymbol{x}_0))_{H_i \times W_i \times c_{in}}$ for a fixed input image $\boldsymbol{x}_0$ (shape: $32 \times 32 \times 3$). The hyperplanes returned from $RM_2$ of Algorithm 1 are $\{\boldsymbol{H}^{Adj}_{s,i}(\boldsymbol{z}(\boldsymbol{x}_0)) \mid s \in \{0, 1, ..., (H_i \times W_i - 1)\}, i \in \{0, 1, ..., (c_{out} - 1)\}\}$. Then we verify that there exists a $\boldsymbol{z}(\boldsymbol{x}_0)$ for our VGG7 or Fixup-ResNet20, such that,

$$c_{s,i} = ((\boldsymbol{F}(\boldsymbol{x}_0))_{H_i \times W_i \times c_{in}} \circledast \boldsymbol{w}_{i, r_1 \times r_2 \times c_{in}})_s \approx \langle \boldsymbol{x}_0 \mid \boldsymbol{H}^{Adj}_{s,i}(\boldsymbol{z}(\boldsymbol{x}_0)) \rangle, \tag{17}$$

where $()_s$ denotes taking a real value from $()$ at offset $s$. In other words, we verify that, with a proper $\boldsymbol{z}$, a unit (offset by $s$) of out-channel feature map (left-hand side of "$\approx$"), is approximate to a dot product between the input image $\boldsymbol{x}_0$ and the effective hyperplane (right-hand side of "$\approx$"), $\boldsymbol{H}^{Adj}_{s,i}(\boldsymbol{z}(\boldsymbol{x}_0))$. Verifying $RM_2$ is equivalent to verifying $RM_4$, $RM_3$, $RM_1$.

**Verify the existence of $\boldsymbol{z}$ on the FC layer:** Suppose $c_k$ denotes an activational value for the class index $k$ of the FC layer, i.e. $c_k = \boldsymbol{F}_k(\boldsymbol{x}_0)$. Also, Algorithm 1 works on the weights of the FC layer with $RM_0$ to generate a hyperplane, $\boldsymbol{H}^{Adj}_k(\boldsymbol{z}(\boldsymbol{x}_0))$. Then according to Eq.10 and 4, we verify that,

$$c_k \approx \langle \boldsymbol{x}_0 \mid \boldsymbol{H}^{Adj}_k(\boldsymbol{z}(\boldsymbol{x}_0)) \rangle. \tag{18}$$

In other words, we verify that a CNN's activation value for the $k^{th}$ class of the FC layer is equivalent to a dot product between the image $\boldsymbol{x}_0$ and the effective hyperplane $\boldsymbol{H}^{Adj}_k(\boldsymbol{z}(\boldsymbol{x}_0))$.

**Experiment:** We validate the above on a 10k test set of CIFAR10 with $\boldsymbol{z}(\boldsymbol{x}) = \frac{\boldsymbol{x}}{8}$ . We calculate the relative errors between feature map neurons (or 10 output neurons of the FC layer), $\boldsymbol{f}_i$, and the approximation $\boldsymbol{p}_i$ based on the dot product (eq.17 or 18), as

$$\boldsymbol{e}_i = \frac{\boldsymbol{p}_i - \boldsymbol{f}_i}{\boldsymbol{f}_{i,\neq 0}}, \tag{19}$$

where $i \in \{0, 1, ..., (c_{out} - 1)\}$ (or $i \in \{0, 1, ..., 9\}$) denotes the $i^{th}$ out-channel feature map (or the $i^{th}$ unit value of the FC layer before ReLU6); $\boldsymbol{f}_{i,\neq 0}$ substitutes all zeros inside $\boldsymbol{f}_i$ with the smallest positive number of the single-precision floating-point (32-bit float, FP32 for short) data type to avoid any divide by zero exception.

**Results:** We illustrate six statistical histograms of relative errors for 5 conv layers and 1 FC layer of VGG7 as figure 7 (Fixup-ResNet20: 19 histograms = 18 conv layers + 1 FC layer, figure 8). Each sample contributes 6 sets (19 sets) of relative errors. The quantity of entries (dimentions of $\boldsymbol{e}_i$) in each set is listed in the $3^{rd}$ column of table 1 (table 3). We collect relative errors calculated from different test samples and split them into their corresponding sets. Then we count the statistics on each set. Summarily, histograms verify that $\exists \boldsymbol{z}$, $\boldsymbol{z} = \frac{\boldsymbol{x}}{8}$, such that $\boldsymbol{H}^{Adj}_{s,i}(\boldsymbol{z}(\boldsymbol{x}))$ and $\boldsymbol{H}^{Adj}_k(\boldsymbol{z}(\boldsymbol{x}))$ achieve Eq.3 with over 99.99% of the relative errors being $\leq 0.01$ in both convolutional layers and the FC layer. Therefore it reveals that AdjointBackMap successfully reconstructs effective hypersurfaces that precisely decide either the feature map or the FC output. Actually, these relative errors are negligible when considering the inaccuracy of the FP32 data type.

Thus we conclude our model satisfies the two necessary conditions with $\boldsymbol{z}(\boldsymbol{x}) = \frac{\boldsymbol{x}}{8}$.

## C.2 Visualization of $\text{RM}_4$

Visualizing $\boldsymbol{H}^{Adj}(\boldsymbol{z}(\boldsymbol{x}_0))$ from $RM_4$ is the most challenging task. The challenge mainly comes from resolution limitation because we decompose every kernel on every stride shift for reconstruction that generates large quantities of effective hyperplanes for visualization. In detail, AdjointBackMap on the Conv1 layer of VGG7 could produce a series of $32 \times 32 \times 3$ shaped $\boldsymbol{H}^{Adj}_{s,j,i}(\boldsymbol{z}(\boldsymbol{x}_0))$ where $s \in \{0, 1, ..., 1023\}, j \in \{0, 1, ..., 31\}, i \in \{0, 1, ..., 31\}$ (Eq.6). If we chose to illustrate them together, the resolution would reach $32,768 \times 32,768 \times 3$, which pushes the plot's DPI over 33k. So we figure all $s$ together in one picture for an in and out channel pair $(j, i)$, named $\boldsymbol{H}^{Adj}_{\{s\},j,i}(\boldsymbol{z}(\boldsymbol{x}_0))$. Preprocessing includes: transpose it by $[3, 0, 4, 1, 2]$, then reshape it to $1024 \times 1024 \times 3$, i.e.,

$$\boldsymbol{H}^{Adj}_{\{s\},j,i}(\boldsymbol{z}(\boldsymbol{x}_0)) = (reshape)(\{\boldsymbol{H}^{Adj}_{s,j,i}(\boldsymbol{z}(\boldsymbol{x}_0)) \mid s \in S\})^T, \tag{20}$$

where $S$ denotes the range of stride offset. We normalize $\boldsymbol{H}^{Adj}_{\{s\},j,i}(\boldsymbol{z}(\boldsymbol{x}_0))$ by max absolute values along input axes of $32 \times 32 \times 3$. Then we take the absolute value on $\boldsymbol{H}^{Adj}_{\{s\},j,i}(\boldsymbol{z}(\boldsymbol{x}_0))$ for visualization. We set the plot DPI with 1k to depict each pixel, which keeps the picture clear and small-sized at the same time. Similar to these, table 2, 4 list dimension information of $\boldsymbol{H}^{Adj}(\boldsymbol{z}(\boldsymbol{x}_0))$ for different layers and different RMs.

**Results:** Two typical images (ship, frog) are selected for experimental input. For every input image, we illustrate five typical patterns for five convolutional layers of VGG7 and four typical patterns for four conv layers selected from four Residual Blocks of Fixup-ResNet20, respectively. Results are illustrated in figure 9 and 10.

### C.3 Visualization of RM$_3$

Dimension of $\boldsymbol{H}^{Adj}(\boldsymbol{z}(\boldsymbol{x}_0))$ by $RM_3$ is reduced as merging happens along with strides. Therefore, we figure $\{\boldsymbol{H}^{Adj}_{j,i}(\boldsymbol{z}(\boldsymbol{x}_0)) \mid j \in \{0, 1, ..., (c_{in} - 1)\}, i \in \{0, 1, ..., (c_{out} - 1)\}\}$ (Eq.7) of a layer together, name as $\boldsymbol{H}^{Adj}_{\{j,i\}}(\boldsymbol{z}(\boldsymbol{x}_0))$. That is,

$$\boldsymbol{H}^{Adj}_{\{j,i\}}(\boldsymbol{z}(\boldsymbol{x}_0)) = (reshape)(\{\boldsymbol{H}^{Adj}_{j,i}(\boldsymbol{z}(\boldsymbol{x}_0))\})^T. \tag{21}$$

We use transpose, reshape, normalization, and absolute value, the same as $RM_4$ for visualization. Every layer has only one picture.

**Results:** We use the two images from $RM_4$. Results are illustrated in figure 11 and 12.

### C.4 Visualization of RM$_2$

$RM_2$ merges in-channel-wise while keeping the stride separated to reconstruct $\{\boldsymbol{H}^{Adj}_{s,i}(\boldsymbol{z}(\boldsymbol{x}_0)) \mid s \in S, i \in \{0, 1, ..., (c_{out} - 1)\}\}$ (Eq.8). As we mentioned, $\boldsymbol{H}^{Adj}_{s,i}(\boldsymbol{z}(\boldsymbol{x}_0))$ from a layer represents an effective hyperplane responsible for a unit in an out-channel feature map. Similar to $RM_4$, resolution restriction forces to figure $\boldsymbol{H}^{Adj}_{s,i}(\boldsymbol{z}(\boldsymbol{x}_0))$ with all stride offset $s$ together in one picture for an out channel $i$, name $\boldsymbol{H}^{Adj}_{\{s\},i}(\boldsymbol{z}(\boldsymbol{x}_0))$. That is,

$$\boldsymbol{H}^{Adj}_{\{s\},i}(\boldsymbol{z}(\boldsymbol{x}_0)) = (reshape)(\{\boldsymbol{H}^{Adj}_{\{s\},i}(\boldsymbol{z}(\boldsymbol{x}_0)) \mid s \in S\})^T. \tag{22}$$

Therefore the quantity of pictures for a convolutional layer is equal to the quantity of the out channels. We apply the same visualization techniques as $RM_4$.

**Results:** We still use the images from $RM_4$. Results are illustrated in figure 13 and 14.

### C.5 Visualization of RM$_1$

$RM_1$ merges both in-channel-wise and stride-wise to generate $\{\boldsymbol{H}^{Adj}_{i}(\boldsymbol{z}(\boldsymbol{x}_0)), i \in \{0, 1, ..., (c_{out}-1)\}\}$ (Eq.9). A $\boldsymbol{H}^{Adj}_{i}(\boldsymbol{z}(\boldsymbol{x}_0))$ on a layer represents an effective hyperplane for a summation of that out-channel feature map.

**Results:** Results are illustrated in figure 15 and 16.

## Appendix D Figures and Tables

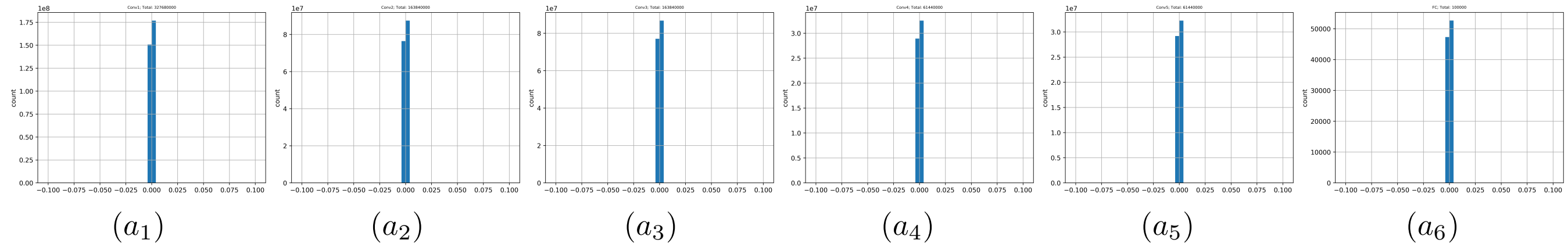

Figure 7: Histogram of relative errors between $\boldsymbol{F}(\boldsymbol{x})$ and $\langle \boldsymbol{x} \mid \boldsymbol{H}^{Adj}(\frac{\boldsymbol{x}}{8})\rangle$, by Conv1$\sim$5 layers and the FC layer (before ReLU6) of VGG7 (eq.17, 18). 10k test samples. The x-axis is the error, and the y-axis is the frequency. According to the $3^{rd}$ column of table 1, a sample generates six sets of relative errors concerning six layers. Here one figure represents statistics of one set collected from the 10k samples. Therefore ($a_1$) to ($a_6$) has quantities of relative errors: $32,768 \times 10k = 327.68m$, $16,384 \times 10k = 163.84m$, $163.84m$, $6,144 \times 10k = 61.44m$, $61.44m$, $100k$, respectively. These quantities have been numerically verified and printed to figure titles ("Total: "). ($a_1$) to ($a_6$) reports the percentage of relative errors (eq.19 $\leq 1\%$): 99.9996%, 99.9996%, 99.9996%, 99.9989%, 99.9989%, 99.996%, respectively.

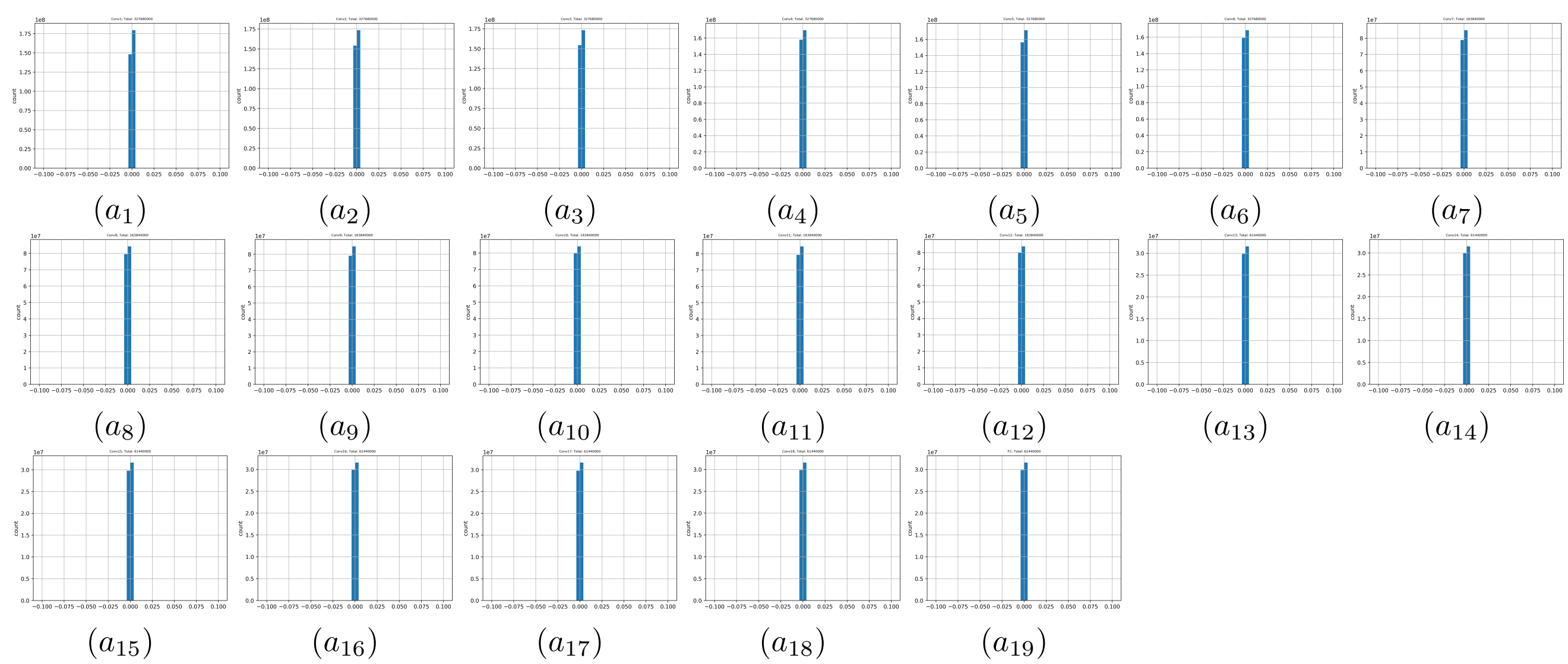

Figure 8: Histogram of relative errors between $\boldsymbol{F}(\boldsymbol{x})$ and $\langle \boldsymbol{x} \mid \boldsymbol{H}^{Adj}(\frac{\boldsymbol{x}}{8})\rangle$ for $i \in \{0, 1, ..., (c_{out} - 1)\}$, by Conv1$\sim$18 layers and the FC layer (before ReLU19) of Fixup-ResNet20. Similar to figure 7 (table 3), ($a_1$) to ($a_{19}$) has quantities of relative errors: $32,768 \times 10k = 327.68m$, $327.68m$, $327.68m$, $327.68m$, $327.68m$, $327.68m$, $16,384 \times 10k = 163.84m$, $163.84m$, $163.84m$, $163.84m$, $163.84m$, $163.84m$, $6,144 \times 10k = 61.44m$, $61.44m$, $61.44m$, $61.44m$, $61.44m$, $61.44m$, $100k$, respectively; These quantities have been numerically verified and printed to figure titles ("Total: "); ($a_1$) to ($a_{19}$) reports all percentages of relative errors (eq.19 $\leq 1\%$) are greater or equal than 99.997%.

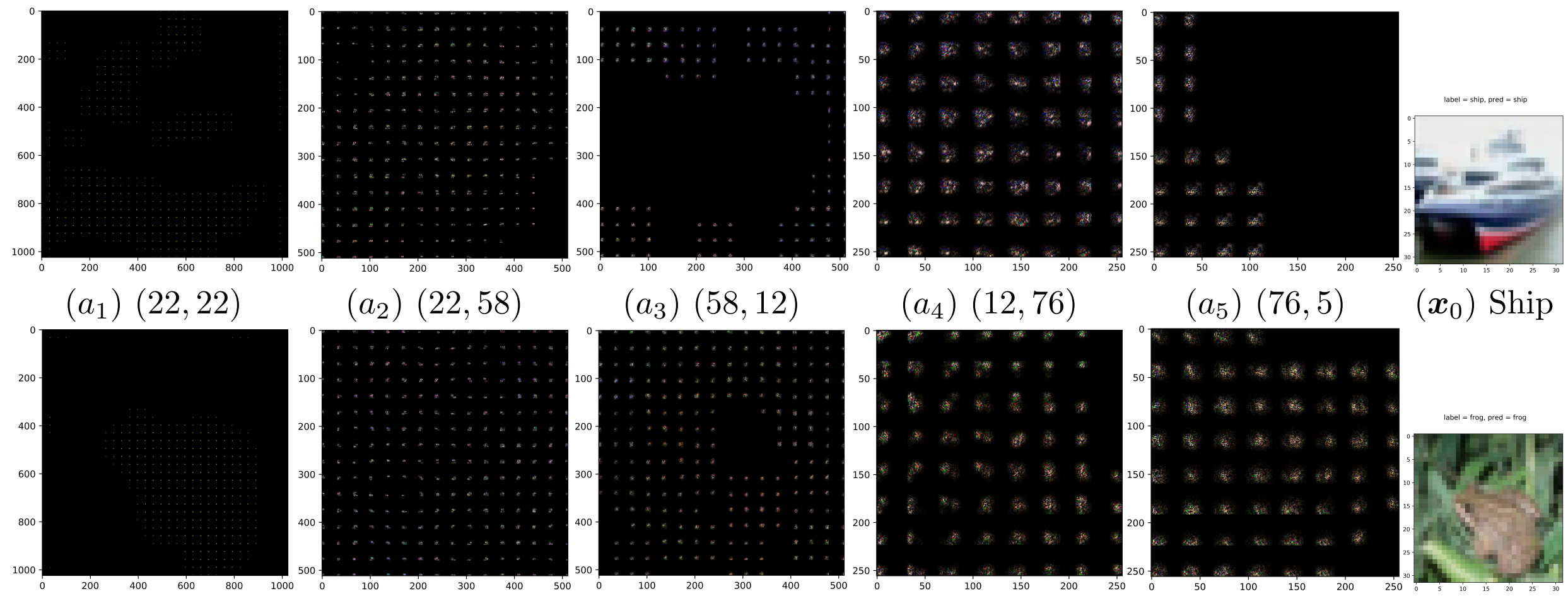

$(a_1)$ $(22, 22)$ $(a_2)$ $(22, 58)$ $(a_3)$ $(58, 12)$ $(a_4)$ $(12, 76)$ $(a_5)$ $(76, 5)$ $(\boldsymbol{x}_0)$ Ship

$(b_1)$ $(22, 22)$ $(b_2)$ $(22, 58)$ $(b_3)$ $(58, 12)$ $(b_4)$ $(12, 76)$ $(b_5)$ $(76, 5)$ $(\boldsymbol{x}_0)$ Frog

Figure 9: Visualization of $\mathbf{RM_4}$ on VGG7. Typical $\boldsymbol{H}^{Adj}_{\{s\},j,i}(\frac{\boldsymbol{x}_0}{8})$ patterns ($a_1$ to $a_5$) mapped from Conv1∼5 layer (Eq.6) by $RM_4$ (eq.20) with a ship input ($b_1$ to $b_5$ with a frog input). $(j, i)$ denotes where a kernel is mapped. Each figure has its shape (quantity of stride moves): $1024 \times 1024 \times 3(32 \times 32), 512 \times 512 \times 3(16 \times 16), 512 \times 512 \times 3(16 \times 16), 256 \times 256 \times 3(8 \times 8), 256 \times 256 \times 3(8 \times 8)$. We see local sparsity (black-out area) increases as layer index decreases because a higher-layer kernel has a larger effective receptive field from the input space perspective.

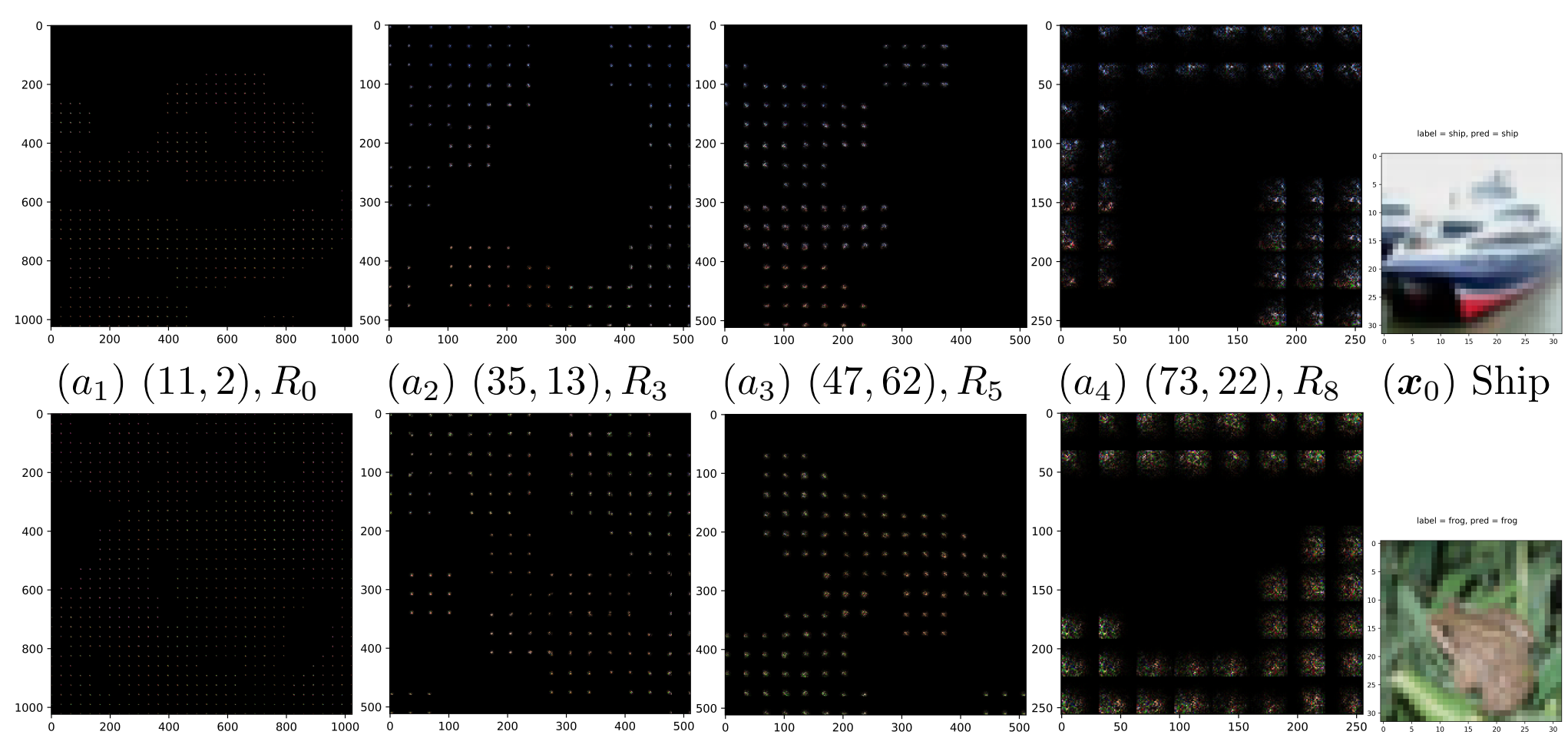

$(a_1)$ $(11, 2), R_0$ $(a_2)$ $(35, 13), R_3$ $(a_3)$ $(47, 62), R_5$ $(a_4)$ $(73, 22), R_8$ $(\boldsymbol{x}_0)$ Ship

$(b_1)$ $(11, 2), R_0$ $(b_2)$ $(35, 13), R_3$ $(b_3)$ $(47, 62), R_5$ $(b_4)$ $(73, 22), R_8$ $(\boldsymbol{x}_0)$ Frog

Figure 10: Visualization of $\mathbf{RM_4}$ on Fixup-ResNet20. Typical $\boldsymbol{H}^{Adj}_{\{s\},j,i}(\frac{\boldsymbol{x}_0}{8})$ patterns ($a_1$ to $a_4$) mapped from the second conv layer (conv$2, 8, 12, 18$) of selected Residual Block ($R$) $0, 3, 5, 8$ (Eq.6) by $RM_4$ (eq.20) with a ship input ($b_1$ to $b_4$ with a frog input). $(j, i)$ denotes where a kernel is mapped. Each figure has its shape (quantity of stride moves): $1024 \times 1024 \times 3(32 \times 32), 512 \times 512 \times 3(16 \times 16), 512 \times 512 \times 3(16 \times 16), 256 \times 256 \times 3(8 \times 8)$. We see local sparsity (black-out area) increases as layer index decreases, similar to figure 9.

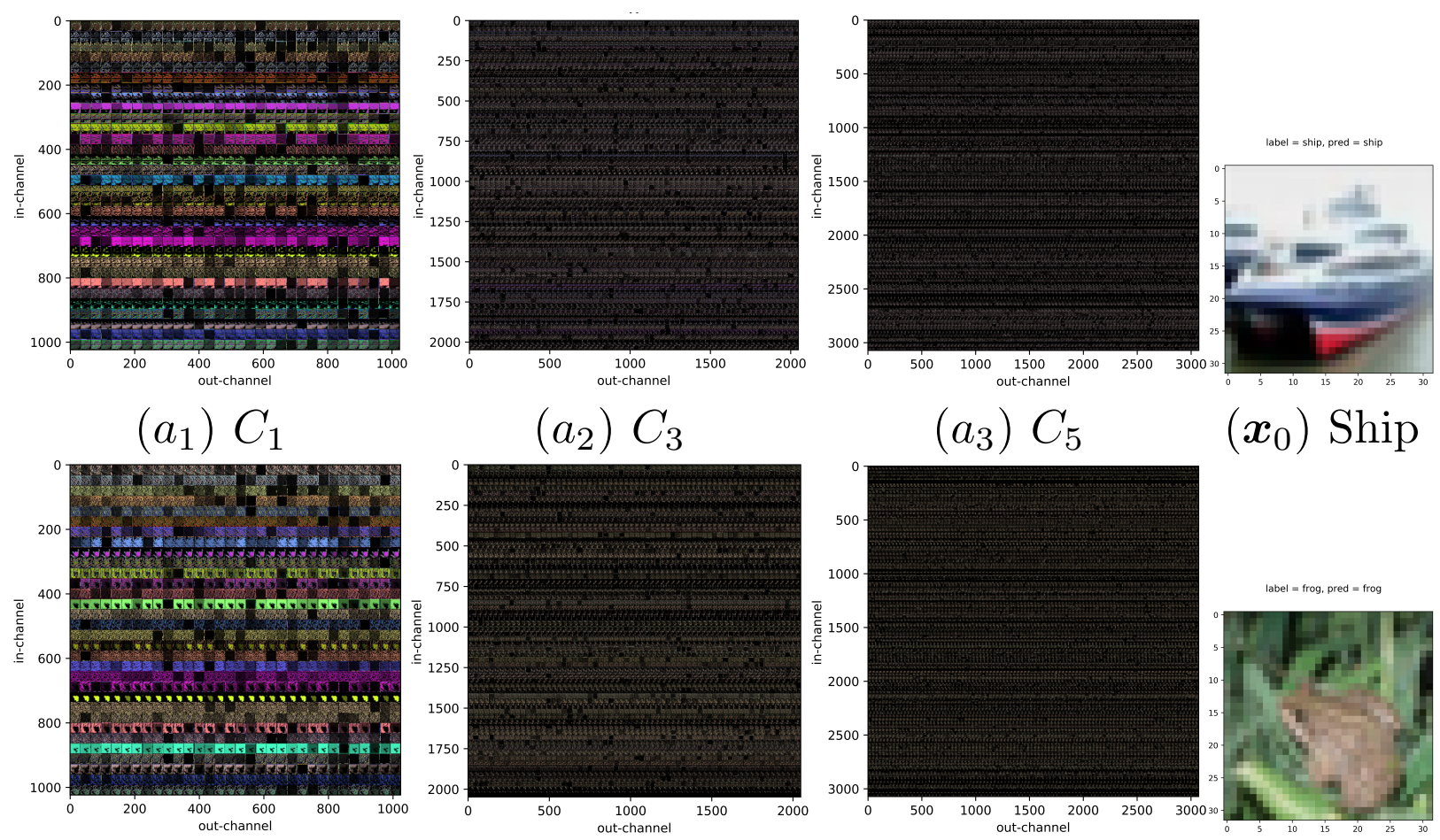

$(a_1)$ $C_1$ $(a_2)$ $C_3$ $(a_3)$ $C_5$ $(\boldsymbol{x}_0)$ Ship

$(b_1)$ $C_1$ $(b_2)$ $C_3$ $(b_3)$ $C_5$ $(\boldsymbol{x}_0)$ Frog

Figure 11: Visualization of $\mathbf{RM_3}$ on VGG7. Typical $\boldsymbol{H}^{Adj}_{\{j,i\}}(\frac{\boldsymbol{x}_0}{8})$ patterns ($a_1$ to $a_3$) mapped from Conv1, 3, 5 ($C_1, C3, C_5$) layer (Eq.7) by $RM_3$ (eq.21) with a ship input ($b_1$ to $b_3$ with a frog input). Each figure has its shape ($c_{in} \times c_{out}$)): $1024 \times 1024 \times 3(32 \times 32), 2048 \times 2048 \times 3(64 \times 64), 3072 \times 3072 \times 3(96 \times 96)$.

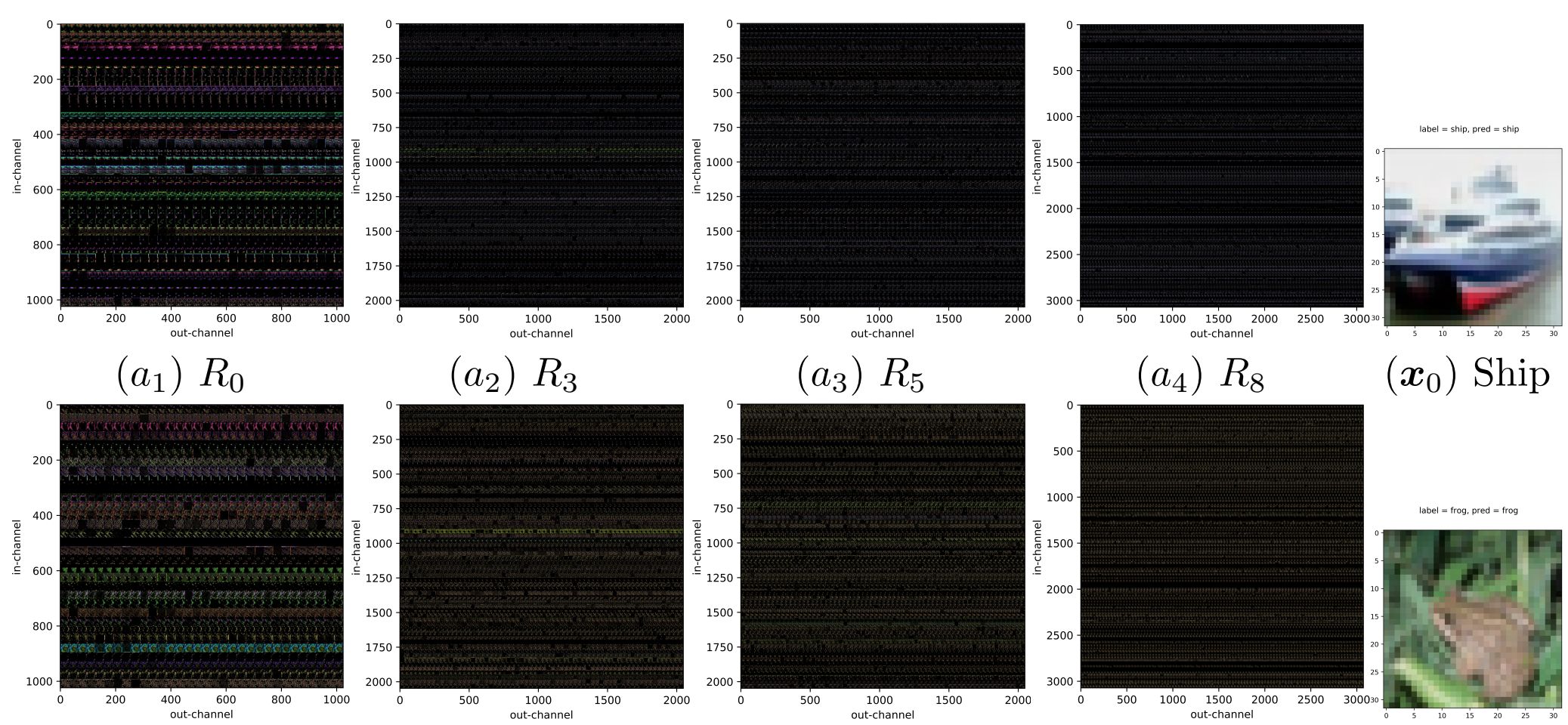

$(a_1)$ $R_0$ $(a_2)$ $R_3$ $(a_3)$ $R_5$ $(a_4)$ $R_8$ $(\boldsymbol{x}_0)$ Ship

$(b_1)$ $R_0$ $(b_2)$ $R_3$ $(b_3)$ $R_5$ $(b_4)$ $R_8$ $(\boldsymbol{x}_0)$ Frog

Figure 12: Visualization of $\mathbf{RM_3}$ on Fixup-ResNet20. Typical $\boldsymbol{H}^{Adj}_{\{j,i\}}(\frac{\boldsymbol{x}_0}{8})$ patterns ($a_1$ to $a_4$) mapped from the second conv layer(conv$2, 8, 12, 18$) of Residual Block ($R$) $0, 3, 5, 8$ (Eq.7) by $RM_3$ (eq.21) with a ship input ($b_1$ to $b_4$ with a frog input). Each figure has its shape ($c_{in} \times c_{out}$)): $1024 \times 1024 \times 3(32 \times 32), 2048 \times 2048 \times 3(64 \times 64), 2048 \times 2048 \times 3(64 \times 64), 3072 \times 3072 \times 3(96 \times 96)$.

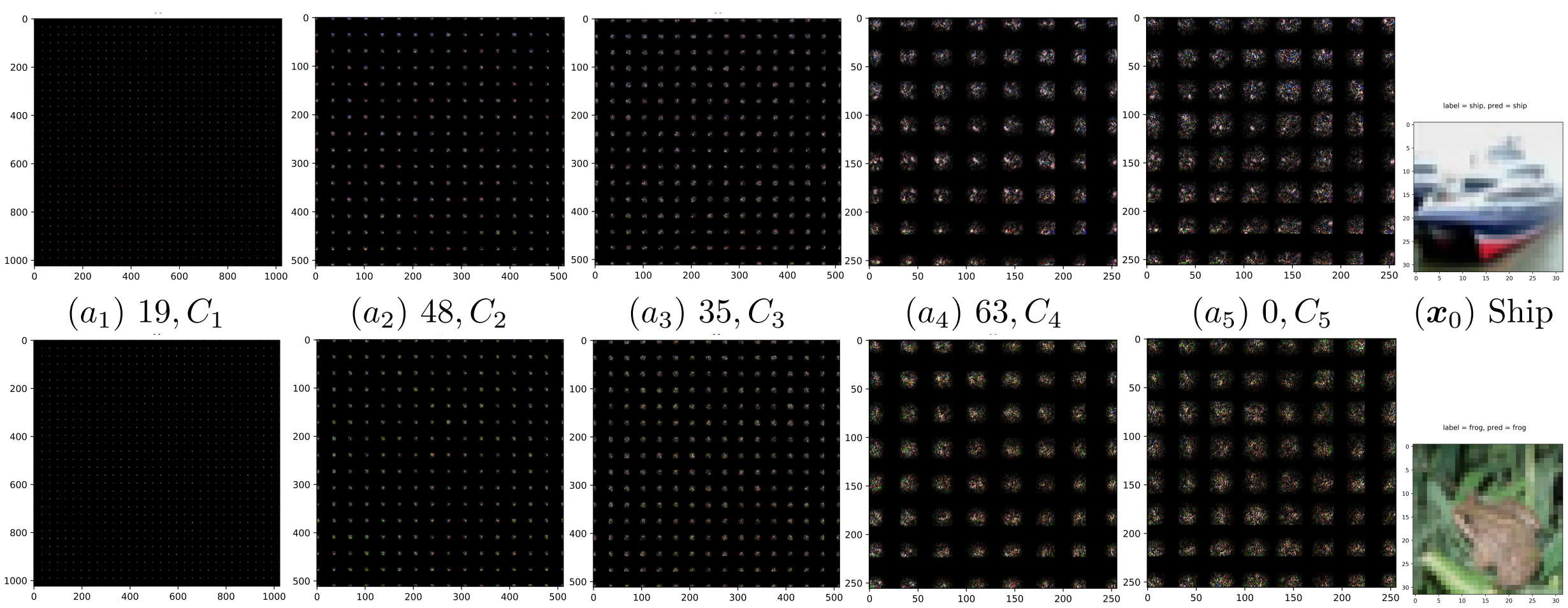

$(a_1)$ $19, C_1$ $(a_2)$ $48, C_2$ $(a_3)$ $35, C_3$ $(a_4)$ $63, C_4$ $(a_5)$ $0, C_5$ $(\boldsymbol{x}_0)$ Ship

$(b_1)$ $19, C_1$ $(b_2)$ $48, C_2$ $(b_3)$ $35, C_3$ $(b_4)$ $63, C_4$ $(b_5)$ $0, C_5$ $(\boldsymbol{x}_0)$ Frog

Figure 13: Visualization of $\mathbf{RM_2}$ on VGG7. Typical $\boldsymbol{H}^{Adj}_{\{s\},i}(\frac{\boldsymbol{x}_0}{8})$ patterns ($a_1$ to $a_5$) mapped from Conv1∼5 ($C_1 \sim C_5$) layer (Eq.8) by $RM_2$ (eq.22) with a ship input ($b_1$ to $b_5$ with a frog input). $i$ denotes the out-channel index where kernels are mapped. Each figure has its shape (quantity of stride moves): $1024 \times 1024 \times 3(32 \times 32), 512 \times 512 \times 3(16 \times 16), 512 \times 512 \times 3(16 \times 16), 256 \times 256 \times 3(8 \times 8), 256 \times 256 \times 3(8 \times 8)$, respectively. Similar to figure 9, a higher-layer kernel has a larger effective receptive field from the input space perspective; Local sparsity (black-out area) will increase when layer index decreases.

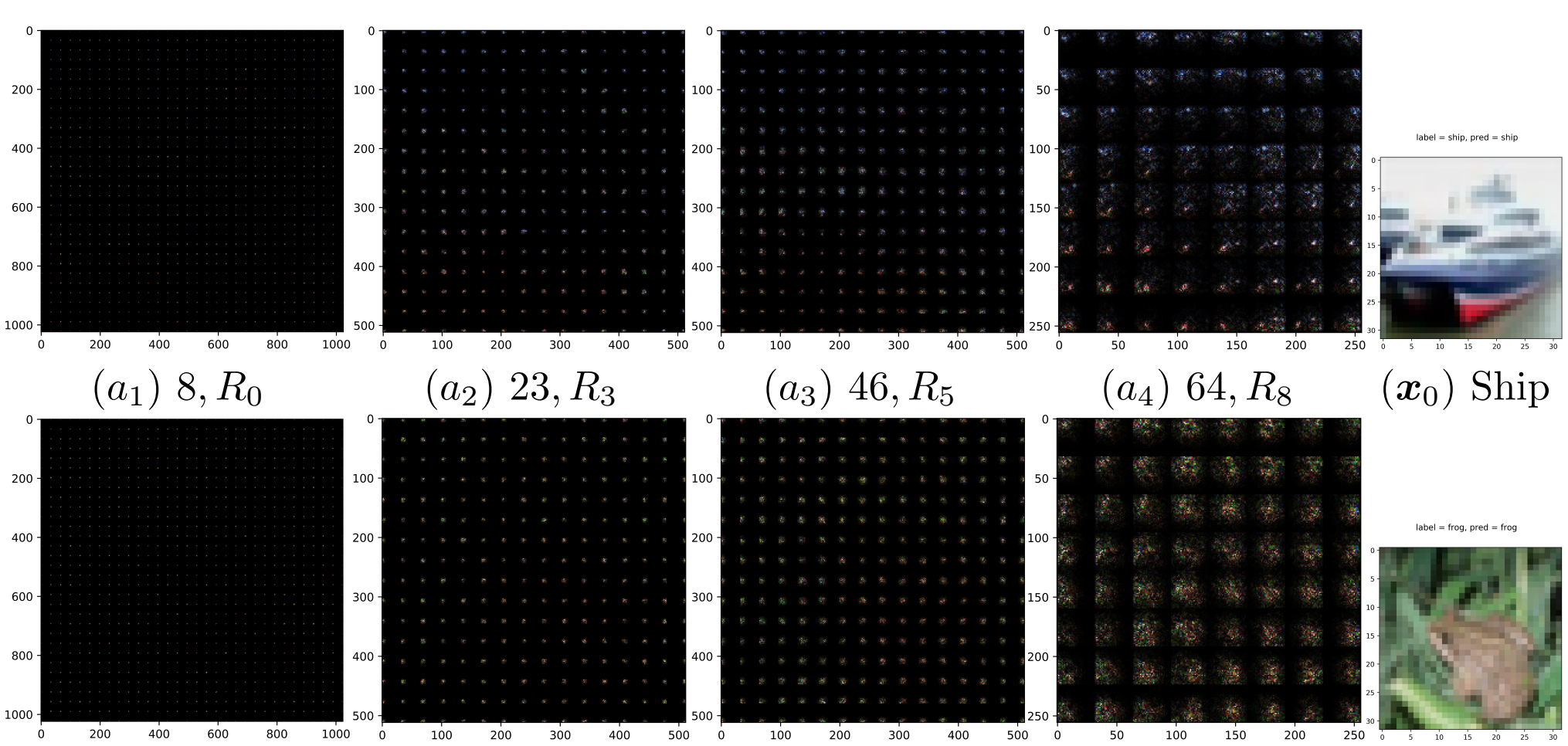

$(a_1)$ $8, R_0$ $(a_2)$ $23, R_3$ $(a_3)$ $46, R_5$ $(a_4)$ $64, R_8$ $(\boldsymbol{x}_0)$ Ship

$(b_1)$ $8, R_0$ $(b_2)$ $23, R_3$ $(b_3)$ $46, R_5$ $(b_4)$ $64, R_8$ $(\boldsymbol{x}_0)$ Frog

Figure 14: Visualization of $\mathbf{RM_2}$ on Fixup-ResNet20. Typical $\boldsymbol{H}^{Adj}_{\{s\},i}(\frac{\boldsymbol{x}_0}{8})$ patterns ($a_1$ to $a_4$) mapped from the second conv layer(conv2, 8, 12, 18) of Residual Block ($R$) $0, 3, 5, 8$ (Eq.8) by $RM_2$ (eq.22) with a ship input ($b_1$ to $b_4$ with a frog input). $i$ denotes the out-channel index where kernels are mapped. Each figure has its shape (quantity of stride moves): $1024 \times 1024 \times 3(32 \times 32), 512 \times 512 \times 3(16 \times 16), 512 \times 512 \times 3(16 \times 16), 256 \times 256 \times 3(8 \times 8)$, respectively. Similar to figure 10, a higher-layer kernel has a larger effective receptive field from the input space perspective; Local sparsity (black-out area) will increase when layer index decreases.

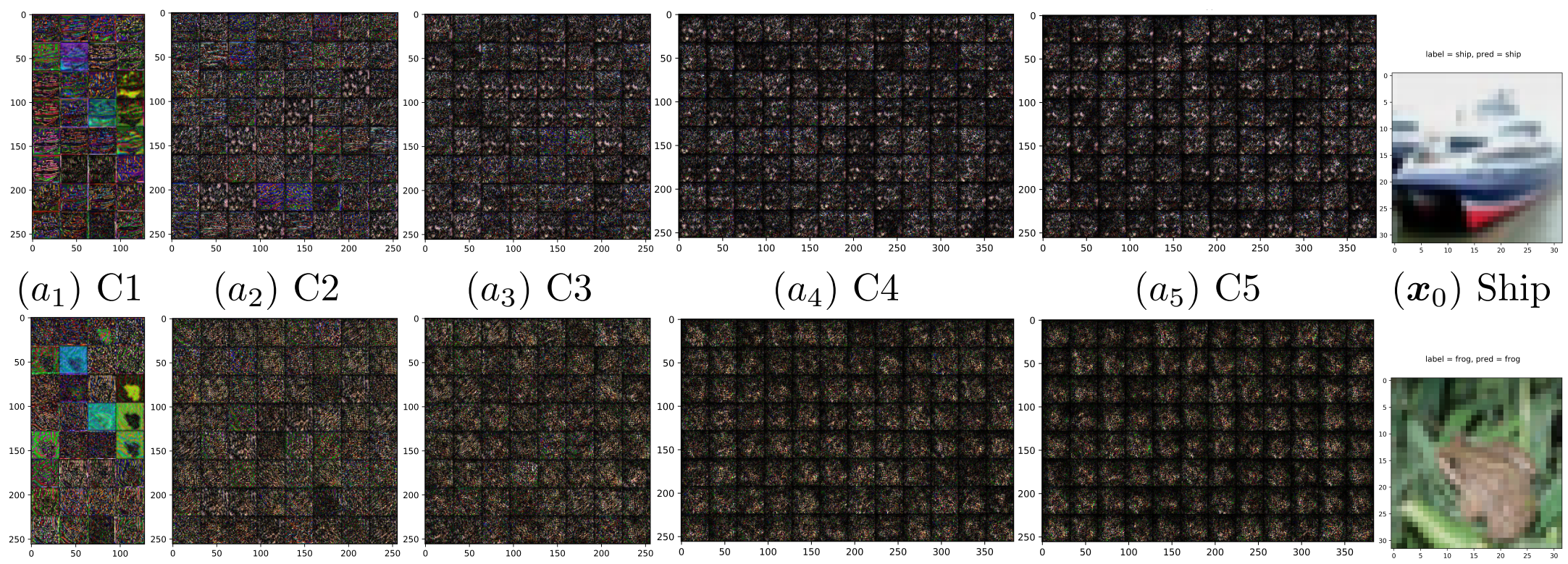

$(a_1)$ C1 $(a_2)$ C2 $(a_3)$ C3 $(a_4)$ C4 $(a_5)$ C5 $(\boldsymbol{x}_0)$ Ship

$(b_1)$ C1 $(b_2)$ C2 $(b_3)$ C3 $(b_4)$ C4 $(b_5)$ C5 $(\boldsymbol{x}_0)$ Frog

Figure 15: Visualization of $\mathbf{RM_1}$ on VGG7. $\boldsymbol{H}_i^{Adj}(\frac{\boldsymbol{x}_0}{8})$ patterns mapped from Conv1∼5 ($C_1 \sim C_5$) layer (Eq.9) by $RM_1$ with a ship input. Quantity of subfigures in a plot is equal to the quantity of out channels.

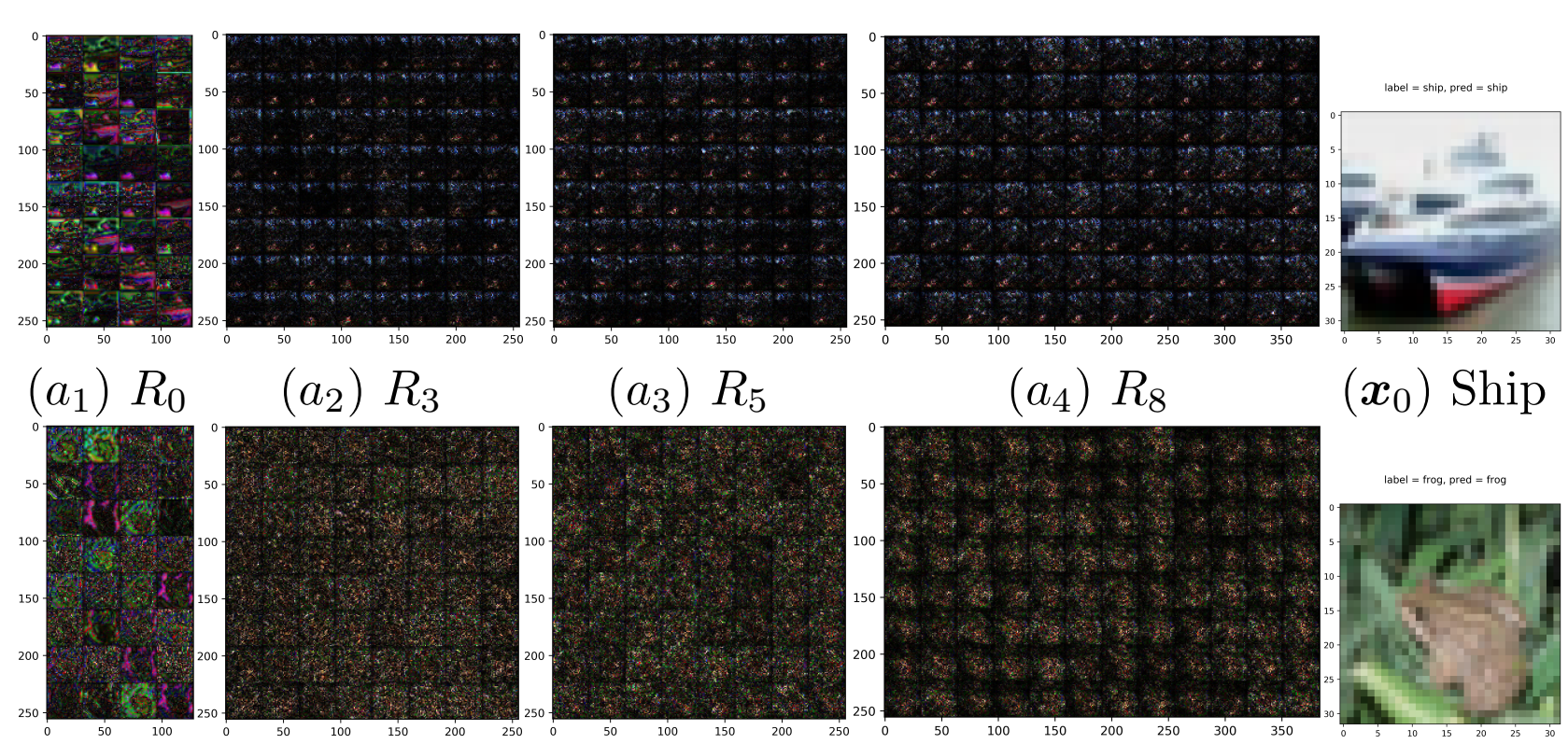

$(a_1)$ $R_0$ $(a_2)$ $R_3$ $(a_3)$ $R_5$ $(a_4)$ $R_8$ $(\boldsymbol{x}_0)$ Ship

$(b_1)$ $R_0$ $(b_2)$ $R_3$ $(b_3)$ $R_5$ $(b_4)$ $R_8$ $(\boldsymbol{x}_0)$ Frog

Figure 16: Visualization of $\mathbf{RM_1}$ on Fixup-ResNet20. $\boldsymbol{H}_i^{Adj}(\frac{\boldsymbol{x}_0}{8})$ patterns ($a_1$ to $a_4$) mapped from the second conv layer(conv2, 8, 12, 18) of Residual Block ($R$) $0, 3, 5, 8$ (Eq.9) by $RM_1$ with a ship input ($b_1$ to $b_4$ with a frog input). Quantity of subfigures in a plot is equal to the quantity of out channels.

| Layer | Parameters | Out-channel Feature Maps |
|---|---|---|
| Conv0 | $3 \times 3 \times 3 \times 32$ | N/S |
| ReLU0 | N/A | N/S |
| Conv1 | $3 \times 3 \times 32 \times 32$ | $32 \times 32 \times 32 (= 32, 768)$ |
| ReLU1 | N/A | N/S |
| Avg-pool-by-2 | N/A | N/S |
| Conv2 | $3 \times 3 \times 32 \times 64$ | $16 \times 16 \times 64 (= 16, 384)$ |
| ReLU2 | N/A | N/S |
| Conv3 | $3 \times 3 \times 64 \times 64$ | $16 \times 16 \times 64 (= 16, 384)$ |
| ReLU3 | N/A | N/S |
| Avg-pool-by-2 | N/A | N/S |
| Conv4 | $3 \times 3 \times 64 \times 96$ | $8 \times 8 \times 96 (= 6, 144)$ |
| ReLU4 | N/A | N/S |
| Conv5 | $3 \times 3 \times 96 \times 96$ | $8 \times 8 \times 96 (= 6, 144)$ |
| ReLU5 | N/A | N/S |
| Global-pool | N/A | N/S |
| FC | [96, 10] | 10 |
| ReLU6 | N/A | N/S |

Table 1: Parameters in VGG7.

Avg-pool-by-2 denotes average pooling with a window size of 3 and stride size of 2;
N/A denotes no learnable parameters;
N/S denotes it is not necessary for our method.

| Layer | $RM_4$ | $RM_3$ | $RM_2$ | $RM_1$ | $RM_0$ |
|---|---|---|---|---|---|
| Conv0 | N/A | N/A | N/A | N/A | N/A |
| Conv1 | $d_{in} \times 32 \times 32 \times 32 \times 32$ | $d_{in} \times 32 \times 32$ | $d_{in} \times 32 \times 32 \times 32$ | $d_{in} \times 32$ | N/A |
| Conv2 | $d_{in} \times 16 \times 16 \times 32 \times 64$ | $d_{in} \times 32 \times 64$ | $d_{in} \times 16 \times 16 \times 64$ | $d_{in} \times 64$ | N/A |
| Conv3 | $d_{in} \times 16 \times 16 \times 64 \times 64$ | $d_{in} \times 64 \times 64$ | $d_{in} \times 16 \times 16 \times 64$ | $d_{in} \times 64$ | N/A |
| Conv4 | $d_{in} \times 8 \times 8 \times 64 \times 96$ | $d_{in} \times 64 \times 96$ | $d_{in} \times 8 \times 8 \times 96$ | $d_{in} \times 96$ | N/A |
| Conv5 | $d_{in} \times 8 \times 8 \times 96 \times 96$ | $d_{in} \times 96 \times 96$ | $d_{in} \times 8 \times 8 \times 96$ | $d_{in} \times 96$ | N/A |
| FC | N/A | N/A | N/A | N/A | $d_{in} \times 10$ |

Table 2: Dimensions of $\boldsymbol{H}^{Adj}(\boldsymbol{z}(\boldsymbol{x}))$ With Different RMs On VGG7.

$d_{in}$ denotes $32 \times 32 \times 3$;
N/A denotes a layer where our AdjointBackMap is not applicable.

| Block (shortcut) | Layer | Parameters | Out-channel Feature Maps |
|---|---|---|---|
| | Conv0 | $3 \times 3 \times 3 \times 32$ | N/S |
| | ReLU0 | N/A | N/S |
| Residual 0 (identity) | Conv1 | $3 \times 3 \times 32 \times 32$ | $32 \times 32 \times 32 (= 32,768)$ |
| | ReLU1 | N/A | N/S |
| | Conv2 | $3 \times 3 \times 32 \times 32$ | $32 \times 32 \times 32 (= 32,768)$ |
| | ReLU2 | N/A | N/S |
| Residual 1 (identity) | Conv3 | $3 \times 3 \times 32 \times 32$ | $32 \times 32 \times 32 (= 32,768)$ |
| | ReLU3 | N/A | N/S |
| | Conv4 | $3 \times 3 \times 32 \times 32$ | $32 \times 32 \times 32 (= 32,768)$ |
| | ReLU4 | N/A | N/S |
| Residual 2 (identity) | Conv5 | $3 \times 3 \times 32 \times 32$ | $32 \times 32 \times 32 (= 32,768)$ |
| | ReLU5 | N/A | N/S |
| | Conv6 | $3 \times 3 \times 32 \times 32$ | $32 \times 32 \times 32 (= 32,768)$ |
| | ReLU6 | N/A | N/S |
| Residual 3 (avg-pool+pad) | Conv7 (stride=2) | $3 \times 3 \times 32 \times 64$ | $16 \times 16 \times 64 (= 16,384)$ |
| | ReLU7 | N/A | N/S |
| | Conv8 | $3 \times 3 \times 64 \times 64$ | $16 \times 16 \times 64 (= 16,384)$ |
| | ReLU8 | N/A | N/S |
| Residual 4 (identity) | Conv9 | $3 \times 3 \times 64 \times 64$ | $16 \times 16 \times 64 (= 16,384)$ |
| | ReLU9 | N/A | N/S |
| | Conv10 | $3 \times 3 \times 64 \times 64$ | $16 \times 16 \times 64 (= 16,384)$ |
| | ReLU10 | N/A | N/S |
| Residual 5 (identity) | Conv11 | $3 \times 3 \times 64 \times 64$ | $16 \times 16 \times 64 (= 16,384)$ |
| | ReLU11 | N/A | N/S |
| | Conv12 | $3 \times 3 \times 64 \times 64$ | $16 \times 16 \times 64 (= 16,384)$ |
| | ReLU12 | N/A | N/S |
| Residual 6 (avg-pool+pad) | Conv13 (stride=2) | $3 \times 3 \times 64 \times 96$ | $8 \times 8 \times 96 (= 6,144)$ |
| | ReLU13 | N/A | N/S |
| | Conv14 | $3 \times 3 \times 96 \times 96$ | $8 \times 8 \times 96 (= 6,144)$ |
| | ReLU14 | N/A | N/S |
| Residual 7 (identity) | Conv15 | $3 \times 3 \times 96 \times 96$ | $8 \times 8 \times 96 (= 6,144)$ |
| | ReLU15 | N/A | N/S |
| | Conv16 | $3 \times 3 \times 96 \times 96$ | $8 \times 8 \times 96 (= 6,144)$ |
| | ReLU16 | N/A | N/S |
| Residual 8 (identity) | Conv17 | $3 \times 3 \times 96 \times 96$ | $8 \times 8 \times 96 (= 6,144)$ |
| | ReLU17 | N/A | N/S |
| | Conv18 | $3 \times 3 \times 96 \times 96$ | $8 \times 8 \times 96 (= 6,144)$ |
| | ReLU18 | N/A | N/S |
| | Global-pool | N/A | N/S |
| | FC | $96 \times 10$ | 10 |
| | ReLU19 | N/A | N/S |

Table 3: Parameters in Fixup-ResNet20 (refer to table 1)

avg-pool+pad denotes average pooling with a window size of 1 and stride size of 2 and then padding zero channels to match the quantity of out channels for summation; weights rescaling is used after Conv2, 4, 6, 8, 10, 12, 14, 16, 18 but not listed here.

| Layer | $RM_4$ | $RM_3$ | $RM_2$ | $RM_1$ | $RM_0$ |
|---|---|---|---|---|---|
| Conv0 | N/A | N/A | N/A | N/A | N/A |
| Conv1 | $d_{in} \times 32 \times 32 \times 32 \times 32$ | $d_{in} \times 32 \times 32$ | $d_{in} \times 32 \times 32 \times 32$ | $d_{in} \times 32$ | N/A |
| Conv2 | $d_{in} \times 32 \times 32 \times 32 \times 32$ | $d_{in} \times 32 \times 32$ | $d_{in} \times 32 \times 32 \times 32$ | $d_{in} \times 32$ | N/A |
| Conv3 | $d_{in} \times 32 \times 32 \times 32 \times 32$ | $d_{in} \times 32 \times 32$ | $d_{in} \times 32 \times 32 \times 32$ | $d_{in} \times 32$ | N/A |
| Conv4 | $d_{in} \times 32 \times 32 \times 32 \times 32$ | $d_{in} \times 32 \times 32$ | $d_{in} \times 32 \times 32 \times 32$ | $d_{in} \times 32$ | N/A |
| Conv5 | $d_{in} \times 32 \times 32 \times 32 \times 32$ | $d_{in} \times 32 \times 32$ | $d_{in} \times 32 \times 32 \times 32$ | $d_{in} \times 32$ | N/A |
| Conv6 | $d_{in} \times 32 \times 32 \times 32 \times 32$ | $d_{in} \times 32 \times 32$ | $d_{in} \times 32 \times 32 \times 32$ | $d_{in} \times 32$ | N/A |
| Conv7 | $d_{in} \times 16 \times 16 \times 32 \times 64$ | $d_{in} \times 32 \times 64$ | $d_{in} \times 16 \times 16 \times 64$ | $d_{in} \times 64$ | N/A |
| Conv8 | $d_{in} \times 16 \times 16 \times 64 \times 64$ | $d_{in} \times 64 \times 64$ | $d_{in} \times 16 \times 16 \times 64$ | $d_{in} \times 64$ | N/A |
| Conv9 | $d_{in} \times 16 \times 16 \times 64 \times 64$ | $d_{in} \times 64 \times 64$ | $d_{in} \times 16 \times 16 \times 64$ | $d_{in} \times 64$ | N/A |
| Conv10 | $d_{in} \times 16 \times 16 \times 64 \times 64$ | $d_{in} \times 64 \times 64$ | $d_{in} \times 16 \times 16 \times 64$ | $d_{in} \times 64$ | N/A |
| Conv11 | $d_{in} \times 16 \times 16 \times 64 \times 64$ | $d_{in} \times 64 \times 64$ | $d_{in} \times 16 \times 16 \times 64$ | $d_{in} \times 64$ | N/A |
| Conv12 | $d_{in} \times 16 \times 16 \times 64 \times 64$ | $d_{in} \times 64 \times 64$ | $d_{in} \times 16 \times 16 \times 64$ | $d_{in} \times 64$ | N/A |
| Conv13 | $d_{in} \times 8 \times 8 \times 64 \times 96$ | $d_{in} \times 64 \times 96$ | $d_{in} \times 8 \times 8 \times 96$ | $d_{in} \times 96$ | N/A |
| Conv14 | $d_{in} \times 8 \times 8 \times 96 \times 96$ | $d_{in} \times 96 \times 96$ | $d_{in} \times 8 \times 8 \times 96$ | $d_{in} \times 96$ | N/A |
| Conv15 | $d_{in} \times 8 \times 8 \times 96 \times 96$ | $d_{in} \times 96 \times 96$ | $d_{in} \times 8 \times 8 \times 96$ | $d_{in} \times 96$ | N/A |
| Conv16 | $d_{in} \times 8 \times 8 \times 96 \times 96$ | $d_{in} \times 96 \times 96$ | $d_{in} \times 8 \times 8 \times 96$ | $d_{in} \times 96$ | N/A |
| Conv17 | $d_{in} \times 8 \times 8 \times 96 \times 96$ | $d_{in} \times 96 \times 96$ | $d_{in} \times 8 \times 8 \times 96$ | $d_{in} \times 96$ | N/A |
| Conv18 | $d_{in} \times 8 \times 8 \times 96 \times 96$ | $d_{in} \times 96 \times 96$ | $d_{in} \times 8 \times 8 \times 96$ | $d_{in} \times 96$ | N/A |
| FC | N/A | N/A | N/A | N/A | $d_{in} \times 10$ |

Table 4: Dimensions of $\boldsymbol{H}^{Adj}(\boldsymbol{z}(\boldsymbol{x}))$ With Different RMs On Fixup-ResNet20.

$d_{in}$ denotes $32 \times 32 \times 3$;
N/A denotes a layer where our AdjointBackMap is not applicable.

| Class index $k$ / Method | $M_1$ | $M_2$ | $M_3$ |
|---|---|---|---|
| 0 (airplane) | 8.31594 | 8.315954 | 8.6690855 |
| 1 (automobile) | 4.0766134 | 4.0766125 | 4.471591 |
| 2 (bird) | 7.1327333 | 7.132732 | 7.3968844 |
| 3 (cat) | 8.730146 | 8.730144 | 8.764506 |
| 4 (deer) | 7.252265 | 7.252265 | 7.06274 |
| 5 (dog) | **9.732837** | **9.732836** | **9.646278** |
| 6 (frog) | 5.3150105 | 5.315008 | 5.101524 |
| 7 (horse) | 9.613704 | 9.613699 | 7.970075 |
| 8 (ship) | −1.3195618 | −1.3195606 | −1.1485753 |
| 9 (truck) | 6.7904973 | 6.7904935 | 7.129147 |

Table 5: Experiment **A**. Verify The Effective Hyperplane (eq.18) For Adding Adversarial Noise to VGG7.

$M_1$ denotes the FC (before ReLU6) output when $(\boldsymbol{x}_0 + \boldsymbol{Advr})$ input;
$M_2$ denotes $\langle \boldsymbol{x}_0 + \boldsymbol{Advr} \mid \boldsymbol{H}_k^{Adj}(\frac{\boldsymbol{x}_0 + \boldsymbol{Advr}}{8}) \rangle$;
$M_3$ denotes $\langle \boldsymbol{x}_0 + \boldsymbol{Advr} \mid \boldsymbol{H}_k^{Adj}(\frac{\boldsymbol{x}_0}{8}) \rangle$;
The third column has significantly smaller errors than the fourth column, which verifies $\boldsymbol{H}_k^{Adj}(\frac{\boldsymbol{x}_0 + \boldsymbol{Advr}}{8})$ (from $M_2$), instead of $\boldsymbol{H}_k^{Adj}(\frac{\boldsymbol{x}_0}{8})$ (from $M_3$), is the effective hyperplane that decides the $k^{th}$ class prediction.

| Class index $k$ / Method | $M_1$ | $M_2$ | $M_3$ |
|---|---|---|---|
| 0 (airplane) | $-0.9798855$ | $-0.97988594$ | 0.0621146 |
| 1 (automobile) | $-2.5764616$ | $-2.5764558$ | $-0.5108745$ |
| 2 (bird) | $-0.5919545$ | $-0.59196144$ | $-6.3072925$ |
| 3 (cat) | 10.356275 | 10.356285 | **7.726749** |
| 4 (deer) | 5.80124 | 5.801249 | $-8.883365$ |
| 5 (dog) | **10.47916** | **10.47917** | $-0.43725738$ |
| 6 (frog) | 3.5252628 | 3.5252705 | $-1.7145596$ |
| 7 (horse) | 6.5016255 | 6.5016294 | $-19.942373$ |
| 8 (ship) | $-6.2591143$ | $-6.2591157$ | $-1.7208233$ |
| 9 (truck) | 2.1569839 | 2.1569812 | $-0.61982846$ |

Table 6: Experiment **A**. Verify The Effective Hyperplane (eq.18) For Adding Adversarial Noise to Fixup-ResNet20. $M_1$ denotes the FC output before ReLU19. Details are similar to table 5 except that the fourth column hits an inaccurate class index (the largest value is "cat" instead of "dog"), which further confirms $\boldsymbol{H}_k^{Adj}(\frac{\boldsymbol{x}_0}{8})$ (from $M_3$) is **NOT** the effective hyperplane for the $k^{th}$ class prediction. As we've verified that $\boldsymbol{H}_k^{Adj}$ is an accurate model (figure 7, 8) for an effective hypersurface reconstruction, both table 5 and 6 further emphasize that a pure accurate model is **not enough**. Only the accurate model working on an accurate input can the decision boundary be precisely reconstructed. This is the point that distinguishes our AdjointBackMap from others (**Related Work** section of the main paper).